\documentclass[journal]{IEEEtran}

\usepackage{graphicx,amsmath, amssymb,lineno}
\usepackage{subfigure, graphicx}
\usepackage{amsmath,amssymb,amsthm}
\usepackage{algorithmic}
\usepackage{algorithm}
\usepackage{tabularx}
\usepackage{multirow}
\usepackage{makecell}

\newtheorem{theorem}{Theorem}
\newtheorem{prop}{Proposition}

\newtheorem{lemma}{Lemma}
\newtheorem{remark}{Remark}
\newtheorem{define}[theorem]{Definition}

\begin{document}

\title{Joint Embedding Learning and Low-Rank Approximation: A Framework for Incomplete Multi-view Learning}

\author{Hong~Tao, Chenping~Hou*,~\IEEEmembership{Member,~IEEE}, Dongyun~Yi, Jubo~Zhu, Dewen~Hu*,~\IEEEmembership{Senior Member,~IEEE}
\thanks{This work was supported by the NSF of China under Grant 61922087 and Grant 61906201, and the NSF for Distinguished Young Scholars of Hunan Province under Grant 2019JJ20020. Chenping Hou and Dewen Hu are the corresponding authors.}
\thanks{Hong Tao, Chenping Hou, Dongyun Yi and Jubo Zhu are with the College of Liberal Arts and Science, National University of Defense Technology, Changsha, 410073, Hunan, China. E-mail: taohong.nudt@hotmail.com, hcpnudt@hotmail.com, dongyun.yi@gmail.com, ju\_bo\_zhu@aliyun.com.}
\thanks{Dewen Hu is with the College of Mechatronics and Automation, National University of Defense Technology, Changsha, 410073, Hunan, China. Email: dwhu@nudt.edu.cn.}
}

\markboth{}%
{Tao \MakeLowercase{\textit{et al.}}: Joint Embedding Learning and Low-Rank Approximation: A Framework for Incomplete Multi-view Learning}

\IEEEtitleabstractindextext{
\begin{abstract}
In real-world applications, not all instances in multi-view data are fully represented.
To deal with incomplete data, Incomplete Multi-view Learning (IML) rises.
In this paper, we propose the Joint Embedding Learning and Low-Rank Approximation (JELLA) framework for IML.
The JELLA framework approximates the incomplete data by a set of low-rank matrices and learns a full and common embedding by linear transformation.
Several existing IML methods can be unified as special cases of the framework.
More interestingly, some linear transformation based complete multi-view methods can be adapted to IML directly with the guidance of the framework.
Thus, the JELLA framework improves the efficiency of processing incomplete multi-view data, and bridges the gap between complete multi-view learning and IML.
Moreover, the JELLA framework can provide guidance for developing new algorithms.
For illustration, within the framework, we propose the Incomplete Multi-view Learning with Block Diagonal Representation (IML-BDR) method.
Assuming that the sampled examples have approximate linear subspace structure, IML-BDR uses the block diagonal structure prior to learn the full embedding, which would lead to more correct clustering.
A convergent alternating iterative algorithm with the Successive Over-Relaxation optimization technique is devised for optimization.
Experimental results on various datasets demonstrate the effectiveness of IML-BDR.

\end{abstract}

}

\maketitle

\IEEEdisplaynontitleabstractindextext

\section{Introduction}
In last decades, multi-view learning has experienced a rapid development, as more and more multi-view data are produced and collected \cite{Luo2013MvML,xu2013MVLsurvey,hou2010MvSSDR,Sun2013A,Zhu2016MvML,Zhan2018Graph,Xue2018LinearCost,TangTMM2018,Zhang2019BinaryMVC,ren2018RAMC}.
Conventional multi-view learning algorithms are mostly developed requiring that each sample is represented fully with all views, i.e., in the complete multi-view setting.
Nevertheless, not all objects can be observed on all views in real-world applications \cite{TangTMM2019MultiLabel,trivedi2010MCIV,Li2014PVC,shao2015MIC, Wang2015PMMH,Shan2018FENH,Wen2019Incomplete,wen2019unified,Zhuge2019SRLC}.
For instance, in video surveillance, the same scene is monitored by multiple cameras from different angles simultaneously, but some cameras could be out of work for some reasons, leading to missing views for some examples (missing-view setting); Moreover, if some cameras work but there are occlusions, then the corresponding variables are vacant in these views.
The mix of missing views and missing variables are regarded as the incomplete-view setting.
In traditional multi-view learning algorithms, there are usually two ways to handle the incomplete multi-view data.
One way is to discard the incomplete examples, which results in losing available information \cite{Zhang2018ICLA}.
The other way is to fill in the missing samples with the mean of the available ones and complete the missing variables by traditional matrix completion algorithms \cite{Li2014PVC,xu2015MVL}.
This saves some useful information, but will still produce inaccuracies.

To handle multi-view data with missing views, the Partial multi-View Clustering (PVC) algorithm is proposed \cite{Li2014PVC}.
Though PVC only deals with the missing-view case, it is a pioneering work of the Incomplete Multi-view Learning (IML).
Concretely, PVC learns a full representation by employing the Nonnegative Matrix Factorization (NMF).
The Multiple Incomplete views Clustering (MIC) \cite{shao2015MIC} algorithm is also based on NMF.
It first fills the missing views by the mean of available examples and then utilizes NMF by allocating smaller weights to the incomplete examples.
To capture the nature of incomplete views and learn a full representation, MIC imposes $\ell_{2,1}$ regularization on each view's feature matrix, and pushes them towards a common consensus.
Based on the assumption that multiple views are generated from a common subspace, the method named Multi-View Learning with Incomplete Views (MVL-IV) is proposed to recover the incomplete instances by multi-view matrix completion \cite{xu2015MVL}.
Recognizing that the previous methods just simply project multiple views to a common subspace, Zhao et al. proposed to incorporate geometric information into the representation and designed the Incomplete Multi-modality Grouping (IMG) method \cite{Zhao2016IMG}.
Specifically, IMG imposes a manifold regularization with automatically learned graph on the common representation to enhance the grouping discriminability.
Yin et al. learned the cluster indicator matrix for incomplete multi-view data directly by preserving both the inter-view and intra-view data similarities in regression \cite{Yin2017UIS}.
To unfold the shared information from different views, Zhang et al. proposed an Isomorphic Linear Correlation Analysis model to learn a feature-isomorphic subspace.
Then, based on the learned feature representation, they utilized an Identical Distribution Pursuit Completion model to complete the missing samples \cite{Zhang2018ICLA}.
Based on weighted semi-nonnegative matrix factorization (semi-NMF), Hu et al. \cite{Hu2018DAIMC} developed the Doubly Aligned Incomplete
Multi-view Clustering (DAIMC) algorithm simultaneously aligning samples and the basis matrices.
Liu et al. \cite{liu2018late} adopted the late fusion strategy to learn a consensus cluster indicator matrix based on the incomplete base cluster indicator matrices produced by incomplete based kernel matrices.

Though being proposed from different perspectives, several existing IML algorithms, e.g., PVC  \cite{Li2014PVC}, MVL-IV \cite{xu2015MVL}, IMG \cite{Zhao2016IMG} and DAIMC \cite{Hu2018DAIMC} are found to have similarities in their forms of formulations.
Hence, in this paper, we propose a general IML framework to reveal their common properties.
The proposed framework performs Joint Embedding Learning and Low-rank Approximation (JELLA), thus we also name it as JELLA.
Concretely, a set of low-rank matrices are introduced to approximate the incomplete representations. The entries of the approximate matrices are constrained to be equal to those of the original data matrices if they are not missing.
Then, the concept of mapping function (e.g., linear transformation) is employed to learn a full and common embedding from multiple views.
That is, the approximate data matrices are mapped to a common representation matrix by using the compatible and complementary information of multiple views.
With this framework, we gain some insight into IML.
On one hand,  some popular IML methods, like PVC, MVL-IV, IMG and DAIMC are unified as special cases of the proposed JELLA framework.
As we will show later, the original two-view PVC and two-view IMG in the missing-view setting can be naturally extended to the multi-view case in the incomplete-view setting.
On the other hand, with the guidance of this framework, some previous multi-view algorithms developed for complete multi-view data, can be adapted to IML directly.
For demonstration, the Robust Multi-view K-Means Clustering (RMKMC) \cite{Cai2013MvKmeans} and Multi-view Concept Learning (MCL) \cite{Guan2015MCL} are adapted to IML in this paper.
Thus, we can deal with the incomplete multi-view data immediately, by taking advantages of the complete multi-view methods, instead of designing a new algorithm.
This bridges the gap between complete multi-view learning and IML, and is of practical significance in improving the efficiency of dealing with incomplete multi-view data.

Within the framework, we propose a specific IML approach for illustration.
Concretely, we assume that the incomplete multi-view data are generated from a union of multiple subspaces.
Note that methods with the block diagonal property would possibly lead to correct subspace clustering \cite{Lu2018BDR}.
To increase the discriminability between groups, we learn the full embedding by Block Diagonal Representation (BDR) \cite{Lu2018BDR}.
That is, the learned full representation matrix is self-expressed with an affinity matrix, on which the block diagonal regularizer is imposed to directly pursue the block diagonal property.
We refer to the proposed method as IML-BDR and adopt the alternating iterative strategy to address it.
A successive over-relaxation (SOR) technique is employed to speed up the convergence.

The contributions are summarized as follows.
\begin{itemize}
\item The JELLA framework is proposed for IML, which includes some popular IML methods as special cases.
    Moreover, with the guidance of the framework, some existing complete multi-view methods can be adapted to IML directly. Thus, JELLA provides a unified perspective for analyzing IML methods, improves the efficiency of processing incomplete multi-view data and bridges the gap between complete multi-view learning and IML.

  \item The JELLA framework provides guidance for designing new IML algorithms. Within the guidance of this framework, we use the block diagonal property to enhance the representation capability of the latent embedding and propose the IML-BDR method.

  \item An algorithm with the successive over-relaxation optimization technique is developed to address the IML-BDR problem, and its convergence is theoretically analyzed.

   \item The effectiveness of IML-BDR is validated by comparing with several state-of-the-art IML methods on various datasets.
\end{itemize}

The structure of the paper is as follows.
Section \ref{sec_RW} introduces the problem setting of IML and briefly reviews some related works.
Section \ref{sec_model} presents the JELLA framework.
The proposed IML-BDR method is introduced in Section \ref{sec_IML_BDR}, and the corresponding experimental results are displayed in Section \ref{sec_exp}.
Finally, we make conclusion in Section \ref{sec_summary}.

\section{Problem Setting and Related Works}\label{sec_RW}
In this section, we first introduce the problem setting of IML. Then, we review several previous works.

Matrices and vectors are denoted by boldface uppercase letters and boldface lowercase letters, e.g., $\mathbf{M}$ and $\mathbf{m}$, respectively.
The $(i,j)$-th entry of $\mathbf{M}$ is denoted as $m_{ij}$ or $\mathbf{M}_{ij}$.
We use $\text{Diag}(\mathbf{m})$ to denote the diagonal matrix with the elements of $\mathbf{m}$ on the main diagonal.
Denote $\text{diag}(\mathbf{M})$ as a vector which is composed of the main diagonal elements of $\mathbf{M} \in \mathbb{R}^{n\times n}$.
The trace, transpose and the Moore-Penrose pseudo-inverse of $\mathbf{M}$ are denoted as $Tr(\mathbf{M})$, $\mathbf{M}^T$ and $\mathbf{M}^{\dagger}$, respectively.
$[\mathbf{M}]_+$ is defined as $ \max(\mathbf{M},0)$.
$\mathbf{M} \geq 0$ means $\mathbf{M}$ is nonnegative. The element-wise product between matrices is denoted by the symbol $\odot$.
The identity matrix, the zero matrix, and the vector with all ones are denoted by $\mathbf{I}$, $\mathbf{0}$ and $\mathbf{1}$, respectively.

The Frobenius norm (or $\ell_2$-norm of a vector) is defined as $\|\mathbf{M}\| = \sqrt{\sum\nolimits_{ij}m_{ij}^2}$ (or $\|\mathbf{m}\| = \sqrt{\sum\nolimits_{i}m_{i}^2}$). The $\ell_{r,p}$-norm ($r\geq 1, p \geq 1$) of $\mathbf{M }\in \mathbb{R}^{d\times n}$ is defined as $\|\mathbf{M}\|_{r,p} = (\sum\limits_{i=1}^d({\sum\limits_{j=1}^n m_{ij}^r})^{\frac{p}{r}})^{\frac{1}{p}}$.
For example, $\|\mathbf{M}\|_{2,1} = \sum\limits_{i=1}^d\sqrt{\sum\limits_{j=1}^n m_{ij}^2}$, $\|\mathbf{M}\|_{1,1} = \sum\limits_{i=1}^d{\sum\limits_{j=1}^n |m_{ij}|}$, $\|\mathbf{M}\|_{\infty,1} = \sum\limits_{i=1}^d{\max\limits_{1\leq j\leq n} |m_{ij}|}$.

\begin{table}
  \centering
  \caption{Notations}\label{tab:notation}
  \begin{tabular}{|l|l|} \hline  
  Notations         & Descriptions                   \\ \hline
  $d$               & The total dimensionality of data \\
  $d^{(v)}$             & The dimensionality of the $v$-th view \\
  $n$               & The data size                        \\
  $k$               & The number of classes                 \\
  $r$               & The low-rank parameter                 \\
  $V$               & The number of views  \\
  $\mathbf{1}$      & A vector of all ones for an arbitrary number  \\
  $\mathbf{I}$   &  The identity matrix \\
  ${{\mathbf{x}}_{i}}\in {{\mathbb{R}}^{d}}$ & The $i$-th data point \\
  ${{\mathbf{x}}_{i}^{(v)}}\in {{\mathbb{R}}^{d^{(v)}}}$ & The $i$-th data points in the $v$-th view \\
  ${\mathbf{X}^{(v)}}\in {{\mathbb{R}}^{d^{(v)}\times n}}$ & The data matrix of the $v$-th view\\
  ${\mathbf{Z}^{(v)}}\in {{\mathbb{R}}^{d^{(v)}\times n}}$ & The approximate data matrix  with rank $r$\\
  $\boldsymbol{\Gamma}^{(v)} \in \{0, 1\}^{d^{(v)} \times n}$ & The $v$-th view's existence-indicating matrix\\
  $\mathbf{W} \in \mathbb{R}^{r \times n}$    & The common representation matrix  \\
  $\mathbf{U}^{(v)}$    & The transformation matrix of the $v$-th view\\
    \multirow{2}*{$\Omega^{(v)}$} & The set of index of available samples  \\
   &  (in the missing-view setting)  \\
\hline 
  \end{tabular}
\end{table}

\subsection{Problem Setting}
Given a data set of $n$ instances $\mathcal{X}=\{\mathbf{x}_i\}_{i=1}^n$, its data matrix is denoted as $\mathbf{X} = [\mathbf{x}_1, \cdots, \mathbf{x}_n] \in \mathbb{R}^{d\times n}$.
Assume that there are $V$ different views, i.e., $\mathbf{x}_i = [\mathbf{x}_i^{(1)}; \cdots; \mathbf{x}_i^{(V)} ]$, where $\mathbf{x}_i^{(v)} \in \mathbb{R}^{d^{(v)}}$ represents the $v$-th view of the $i$-th data point.
Note that $d = \sum\limits_{v=1}^V d^{(v)}$.
The data matrix of the $v$-th view is denoted as $\mathbf{X}^{(v)} = [\mathbf{x}_1^{(v)}, ..., \mathbf{x}_n^{(v)}]$.

The incomplete multi-view setting is defined as the situation that each view lacks some data.
In more detail, one sample could lose its entire representation on a certain view (i.e., missing view) or lose some entries of the data matrix (i.e., missing variables).
That is, some columns of the data matrix $\mathbf{X}^{(v)}$ may be completely or partially vacant.
For representational convenience, an existence-indicating matrix $\boldsymbol{\Gamma}^{(v)} \in \{0, 1\}^{d^{(v)} \times n}$ is used to record the availability of variables in $\mathbf{X}^{(v)}$ ($v = 1, \cdots, V$).
$\boldsymbol{\Gamma}_{ij}^{(v)}$ equals to 1 if and only if $\mathbf{X}_{ij}^{(v)}$ is not missing.
For the missing-view setting,  the set of indexes of available examples on the $v$-th view is notated as $\Omega^{(v)}$, and the corresponding data matrix is $\bar{\mathbf{X}}^{(v)} =  \mathbf{X}_{i:i\in \Omega^{(v)}}^{(v)} \in \mathbb{R}^{d^{(v)}\times n^{(v)}}$, where $n^{(v)} = |\Omega^{(v)}| (\leq n)$ is the number of available examples on the $v$-th view.

Table \ref{tab:notation} lists the descriptions of some frequently used variables in this paper.

\subsection{Partial Multi-view Clustering}
PVC \cite{Li2014PVC} is originally designed for two-view data in the missing view setting. It learns a full representation from incomplete multi-view data based on the NMF.
Denote $\hat{\mathbf{X}}^{(1,2)} = [\mathbf{X}_c^{(1)}; \mathbf{X}_c^{(2)}]$ as the examples presented in both views, and denote $\hat{\mathbf{X}}^{(1)}$, $\hat{\mathbf{X}}^{(2)}$ as the examples only presented in the first view and the second view, respectively. The optimization problems of PVC is\footnote{From codes (http://lamda.nju.edu.cn/code\_PVC.ashx) published by the authors, it can be known that the $\ell_1$-norm in PVC \cite{Li2014PVC} is actually the $\ell_{1,1}$-norm of matrix.}
\begin{equation}\label{eq_PVC}
\begin{array}{l}
\mathop {\min}\limits_{\substack{\mathbf{U}^{(1)}, \mathbf{U}^{(2)},\\ \bar{\mathbf{W}}^{(1)},\bar{\mathbf{W}}^{(2)}}}  \|[\mathbf{X}_c^{(1)}, \hat{\mathbf{X}}^{(1)}] - \mathbf{U}^{(1)}[\mathbf{W}_c, \hat{\mathbf{W}}^{(1)}]\|^2 + \alpha \|\bar{\mathbf{W}}^{(1)}\|_{1,1} \\
\qquad ~~ +   \|[\mathbf{X}_c^{(2)}, \hat{\mathbf{X}}^{(2)}] - \mathbf{U}^{(2)}[\mathbf{W}_c, \hat{\mathbf{W}}^{(2)}]\|^2 + \alpha \|\bar{\mathbf{W}}^{(2)}\|_{1,1} \\
~~  \text{s.t.} ~~  \mathbf{U}^{(v)} \geq 0, \bar{\mathbf{W}}^{(v)} \geq 0, v = 1, 2, \\
\end{array}
\end{equation}
where $\bar{\mathbf{W}}^{(v)} = [\mathbf{W}_c , \hat{\mathbf{W}}^{(v)}] \in \mathbb{R}^{r \times n^{(v)}}$ $(v = 1, 2)$ is the latent representation for the $v$-th view, and $\mathbf{U}^{(v)} \in \mathbb{R}^{d^{(v)} \times r}$ is the basis matrix, $r$ is the dimension of the latent space, and $\alpha > 0$ is the trade-off parameter for the regularization terms.

\subsection{Multi-view Learning with Incomplete Views}
Based on multi-view matrix completion, MVL-IV tends to recover the incomplete multi-view data $\{\mathbf{X}^{(v)}\}_{v=1}^V$ by exploring the connection among multiple views \cite{xu2015MVL}.
Denote the reconstructed data matrices as $\{\mathbf{Z}^{(v)} \in \mathbb{R}^{d^{(v)} \times n} \}_{v=1}^V$, the formulation of  MVL-IV is
\begin{equation}\label{eq_MVL}
\begin{array}{cl}
\mathop {\min}\limits_{\substack{\{\mathbf{U}^{(v)}, \mathbf{Z}^{(v)}\}_{v=1}^V, \\\mathbf{W}}} & \sum\limits_{v = 1}^V \|\mathbf{Z}^{(v)} - \mathbf{U}^{(v)}\mathbf{W}\|^2 \\
\text{s.t.} &\boldsymbol{\Gamma}^{(v)} \odot (\mathbf{Z}^{(v)}-\mathbf{X}^{(v)}) = \mathbf{0},
\end{array}
\end{equation}
where $\mathbf{U}^{(v)} \in \mathbb{R}^{d^{(v)} \times r}$ is the basis matrix, and $\mathbf{W} \in \mathbb{R}^{r \times n}$ is the full representation matrix, and $\odot$ denotes the element-wise product between matrices. MVL-IV is able to cope with the complex incomplete-view setting with both missing views and missing variables.

\subsection{Incomplete Multi-modality Grouping}
The Incomplete Multi-modality Grouping (IMG) approach can be regarded as an enhanced version of PVC.
Differently, IMG gets rid of the nonnegative constraint and considers the global structure in the latent space \cite{Zhao2016IMG}.
Using the same notations with PVC, the latent representation of all samples can be denoted as $\mathbf{W} = [\mathbf{W}_c, \hat{\mathbf{W}}^{(1)}, \hat{\mathbf{W}}^{(2)}]$.
With a Laplacian graph regularization (LGR) to capture the global structure, the objective function of IMG is
\begin{equation}\label{eq_IMG}
\begin{array}{l}
\mathop {\min}\limits_{\substack{\mathbf{U}^{(1)}, \mathbf{U}^{(2)},\mathbf{W},\\ \mathbf{A}\mathbf{1} = \mathbf{1}, \mathbf{A} \geq 0}}  \|[\mathbf{X}_c^{(1)}, \hat{\mathbf{X}}^{(1)}] - \mathbf{U}^{(1)}[\mathbf{W}_c, \hat{\mathbf{W}}^{(1)}]\|^2 + \alpha\|\mathbf{U}^{(1)}\|^2  \\
\qquad \qquad+ \|[\mathbf{X}_c^{(2)}, \hat{\mathbf{X}}^{(2)}] - \mathbf{U}^{(2)}[\mathbf{W}_c, \hat{\mathbf{W}}^{(2)}]\|^2
+ \alpha\|\mathbf{U}^{(2)}\|^2 \\
\qquad \qquad+ \beta Tr(\mathbf{W}\mathbf{L}_{\mathbf{A}}\mathbf{W}^T) + \gamma \|\mathbf{A}\|^2\\
\end{array}
\end{equation}
where $\mathbf{L}_{\mathbf{A}} = \text{Diag}(\mathbf{A}\mathbf{1})- \mathbf{A}$ is the Laplacian matrix of similarity matrix $\mathbf{A}$, $\alpha$, $\beta$ and $\gamma$ are positive parameters.

\subsection{Doubly Aligned Incomplete Multi-view Clustering}
The Doubly Aligned Incomplete Multi-view Clustering (DAIMC) \cite{Hu2018DAIMC} method adopts the weighted semi-nonnegative matrix factorization (semi-NMF) to learn a common latent matrix $\mathbf{W} \in \mathbb{R}^{r \times n}$.
Meanwhile, DAIMC uses the $\ell_{2,1}$-norm regularized regression ($\ell_{2,1}$-RR) to align different basis matrices $\{\mathbf{U}^{(v)} \in \mathbb{R}^{d^{(v)}\times n}\}_{v=1}^V$.
The formulation of DAIMC is
\begin{equation}\label{eq_DAIMC}
\begin{array}{l}
\mathop {\min}\limits_{\substack{\{\mathbf{U}^{(v)},\mathbf{B}^{(v)} \}_{v=1}^V, \\ \mathbf{W} \geq 0}}  \sum\limits_{v=1}^V \big\{||(\mathbf{X}^{(v)} - \mathbf{U}^{(v)}\mathbf{W})\mathbf{P}^{(v)}||^2 \\
\qquad \qquad \qquad + \alpha (\|(\mathbf{B}^{(v)})^T\mathbf{U}^{(v)} - \mathbf{I}\|^2 + \beta \|\mathbf{B}^{(v)}\|_{2,1})\big\},\\
\end{array}
\end{equation}
where $\alpha$ and $\beta$ are nonnegative trade-off parameters,  $\mathbf{I}$ is the identity matrix, and $\mathbf{P}^{(v)}$ is a diagonal matrix. $\mathbf{P}_{ii}^{(v)} = 1$ if the $i$-th instance is in the $v$-th view, otherwise $\mathbf{P}_{ii}^{(v)} = 0$.

\section{IML via Joint Embedding Learning and Low-Rank Approximation}\label{sec_model}
In this section, the formula of the proposed framework and its optimization strategy are presented firstly.
Then, we show some previous IML methods are special cases of the framework. Finally, we show that some complete multi-view methods can be adapted to IML directly with the guidance of the framework.

\subsection{The Formulation}
To complete a matrix with random missing values, one usually uses the low-rank assumption \cite{Zhang2018ICLA,Peng2014lowrank,Zhang2019Hashing,TangTKDE2019FS}.
When dealing with missing views, the low-rank assumption alone is not able to produce satisfactory results \cite{xu2015MVL}.
Fortunately, Xu et al. \cite{xu2015MVL} have shown that the missing views can be restored by low-rank matrices with the help of the connection between
multiple views.
Thus, in the proposed framework, the original incomplete representations $\{\mathbf{X}^{(v)} \}_{v=1}^V$ are approximated by a set of low-rank matrices $\{\mathbf{Z}^{(v)} \}_{v=1}^V$.
The entries of the approximate matrices are constrained to be equal to those of the original data matrices if they are not missing.
To learn a common and full embedding from multiple views, the concept of mapping function is employed.
More concretely, the widely-used linear transformation is employed as the mapping function, due to its convenience in computation and easy-to-explain nature in many applications \cite{long2008general}.
Recall that the low-rank factorization is a special case of linear transformation.
That is, a matrix $\mathbf{Z}^{(v)} \in \mathbb{R}^{d^{(v)} \times n} $ with rank no more than $r$ can be decomposed into the form $\mathbf{Z}^{(v)} = \mathbf{U}^{(v)}\mathbf{W}$, where $\mathbf{U}^{(v)} \in \mathbb{R}^{d^{(v)} \times r}$ and $\mathbf{W} \in \mathbb{R}^{r \times n} $.
Hence, the objective function of the JELLA framework can be formulated as
\begin{equation}\label{eq_general1}
\begin{array}{l}
\mathop {\min}\limits_{\substack{\mathbf{W},\\\{\mathbf{U}^{(v)}, \mathbf{Z}^{(v)}\}}} \!\!\sum\limits_{v = 1}^V \!f^{(v)}(\mathbf{Z}^{(v)},\mathbf{U}^{(v)}\mathbf{W}) \!+ \! \gamma_1 \mathcal{R}_1(\mathbf{U}^{(v)})  \!+ \!\gamma_2 \mathcal{R}_2(\mathbf{W})\\
\quad \text{s.t.} ~~ \boldsymbol{\Gamma}^{(v)} \odot (\mathbf{Z}^{(v)}-\mathbf{X}^{(v)}) = \mathbf{0},
\mathbf{U}^{(v)} \in \mathcal{C}_1^{(v)}, \mathbf{W} \in \mathcal{C}_2, \forall v,\\
\end{array}
\end{equation}
where $f^{(v)}(\mathbf{Z}^{(v)},\mathbf{U}^{(v)}\mathbf{W})$ is the loss function, $\mathbf{Z}^{(v)}$ is the reconstructed low-rank data matrix, $\mathbf{W}$ is the learned common and full embedding matrix, $\mathbf{U}^{(v)}$ is the linear transformation matrix between  $\mathbf{Z}^{(v)}$ and $\mathbf{W}$, and $r \le \min \{d^{(1)}, \cdots, d^{(V)}, n\}$ is a parameter to be determined. $\mathcal{R}_1(\mathbf{U}^{(v)})$ and $\mathcal{R}_2(\mathbf{W})$ are the regularization term on $\mathbf{U}^{(v)}$ and $\mathbf{W}$ with nonnegative parameters $\gamma_1$ and $\gamma_2 $.
The constraint $\boldsymbol{\Gamma}^{(v)} \odot (\mathbf{Z}^{(v)}-\mathbf{X}^{(v)}) = \mathbf{0}$ imposes the entries of $\mathbf{Z}^{(v)}$ to be equal to those of $\mathbf{X}^{(v)}$, at the positions where corresponding variables are not missing.
$\mathcal{C}_1^{(v)}$ and $\mathcal{C}_2$ are constraints on $\mathbf{U}^{(v)}$ and $\mathbf{W}$, respectively.

\begin{remark}
When the original data $\{\mathbf{X}^{(v)} \}_{v=1}^V$ are full, it is unnecessary to introduce the low-rank matrices $\{\mathbf{Z}^{(v)}\}_{v=1}^V$, and the formulation reduces to a complete multi-view model. Thus, the JELLA framework is scalable to deal with full or missing multi-view data, bridging the gap between complete multi-view learning and IML.
\end{remark}

If the incomplete data are all with missing views, i.e., only in the missing-view setting, then the objective function of JELLA can be rewritten as
\begin{equation}\label{eq_general2}
\begin{array}{l}
\mathop {\min}\limits_{\substack{\mathbf{W},\\\{\mathbf{U}^{(v)}\}_{v=1}^V}}\!\!\!\! \sum\limits_{v = 1}^V  \! f^{(v)}(\bar{\mathbf{X}}^{(v)},\mathbf{U}^{(v)}\bar{\mathbf{W}}^{(v)}) \!+ \!\gamma_1 \mathcal{R}_1(\mathbf{U}^{(v)}) \!+ \!\gamma_2 \mathcal{R}_2(\mathbf{W})\\
\quad \text{s.t.} ~~~   \mathbf{U}^{(v)} \in \mathcal{C}_1^{(v)}, \mathbf{W} \in \mathcal{C}_2, \forall v,\\
\end{array}
\end{equation}
where $\bar{\mathbf{X}}^{(v)} =  \mathbf{X}_{i:i\in \Omega^{(v)}}^{(v)} \in \mathbb{R}^{d^{(v)}\times n^{(v)}}$ and $\bar{\mathbf{W}}^{(v)} =  \mathbf{W}_{i:i\in \Omega^{(v)}} \in \mathbb{R}^{r\times n^{(v)}}$ denote the data matrix and latent representation of the survived samples on the $v$-th view, respectively, $\Omega^{(v)}$ is the set of indexes of survived samples on the $v$-th view and $|\Omega^{(v)}|=n^{(v)}$.

\subsection{Optimization Strategy}
Since the resultant formulations usually have multiple groups of unknown variables and the objective are non-convex, it is hard to optimize all unknown variables simultaneously.
Hence, this kind of objectives are often solved by the alternative minimizing strategy. That is, iteratively optimizing one group of variables at a time with the other variables fixed as constants.
Algorithm \ref{alg1} describes the details for solving the general problem (\ref{eq_general1}). If the regularization and constraint on $\mathbf{W}$ are separable, then problem (\ref{eq_general2})\footnote{Sometimes, we may need to introduce an auxiliary variable for $\mathbf{W}$ and optimize the corresponding augmented Lagrangian function.} can be addressed by Algorithm \ref{alg2}.

\begin{algorithm}
\caption{The algorithm to solve problem (\ref{eq_general1})}
\label{alg1}
\begin{algorithmic}
\STATE \textbf{Input:}
$\{\mathbf{X}^{(v)}\}_{v=1}^V$, $\{\mathbf{\Gamma}^{(v)} \}_{v=1}^V$, initial $\{\mathbf{Z}_0^{(v)} \}_{v=1}^V$ and $\mathbf{W}_0$, and nonnegative parameters $\lambda_1$, $\lambda_2$, $t = 0$;

\STATE \textbf{Output:}
$\mathbf{W}$, $\mathbf{U}^{(v)}$, and $\mathbf{Z}^{(v)} (\forall v \in [1, V])$.
\STATE \textbf{while} not converged \textbf{do}
\STATE 1: Update $\mathbf{U}_{t+1}^{(v)}$ ($\forall v \in [1, V]$ ) by solving
\[
\mathop {\min}\limits_{\mathbf{U}^{(v)} \in \mathcal{C}_1^{(v)}} f^{(v)}(\mathbf{Z}_t^{(v)}, \mathbf{U}^{(v)}\mathbf{W}_t) + \lambda_1 \mathcal{R}_1(\mathbf{U}^{(v)}).
\]

\STATE 2: Update $\mathbf{W}_{t+1}$ by solving
\[
\mathop {\min}\limits_{ \mathbf{W} \in \mathcal{C}_2} \sum\limits_{v = 1}^V f^{(v)}(\mathbf{Z}_t^{(v)}, \mathbf{U}_{t+1}^{(v)}\mathbf{W} ) + \lambda_2 \mathcal{R}_2(\mathbf{W}).
\]
\STATE 3: Update $\mathbf{Z}_{t+1}^{(v)}$ by solving
\[
\mathop {\min}\limits_{\boldsymbol{\Gamma}^{(v)}\odot (\mathbf{Z}^{(v)}-\mathbf{X}^{(v)}) = \mathbf{0}}  f^{(v)}(\mathbf{Z}^{(v)},\mathbf{U}_{t+1}^{(v)}\mathbf{W}_{t+1}).
\]
\STATE 4: $t = t+1$.
\STATE \textbf{end while}
\end{algorithmic}
\end{algorithm}

\begin{algorithm}
\caption{The algorithm to solve problem (\ref{eq_general2})}
\label{alg2}
\begin{algorithmic}
\STATE \textbf{Input:}
$\{\bar{\mathbf{X}}^{(v)}\}_{v=1}^V$, $\{{\Omega}^{(v)} \}_{v=1}^V$, initial $\mathbf{W}_0$, and nonnegative parameters $\lambda_1$, $\lambda_2$, $t = 0$;

\STATE \textbf{Output:}
$\mathbf{U}^{(v)} (\forall v \in [1, V])$ and  $\mathbf{W}$.
\STATE \textbf{while} not converged \textbf{do}
\STATE 1: Update $\mathbf{U}_{t+1}^{(v)}$ ($\forall v \in [1, V]$) by solving
\[
\mathop {\min}\limits_{\mathbf{U}^{(v)} \in \mathcal{C}_1^{(v)}} f^{(v)}(\bar{\mathbf{X}}^{(v)}, \mathbf{U}^{(v)}\bar{\mathbf{W}}_t^{(v)} )+ \lambda_1 \mathcal{R}_1(\mathbf{U}^{(v)}).
\]
\STATE 2: Update $\mathbf{w}_{i,t+1}$, $\forall i \in [1, n]$ by solving
\[
\mathop {\min}\limits_{ \mathbf{w}_i \in \mathcal{C}_2} \sum\limits_{v \in \{v'|i\in \Omega^{(v')}\}} f^{(v)}(\mathbf{x}_i^{(v)},\mathbf{U}_{t+1}^{(v)}\mathbf{w}_i) + \lambda_2 \mathcal{R}_2(\mathbf{w}_i),
 \]
$\{v'|i \in \Omega^{(v')}\}$ denotes the set of views where $\mathbf{x}_i$ is not missing.
\STATE 3: $t = t+1$.
\STATE \textbf{end while}
\end{algorithmic}
\end{algorithm}

\begin{table*}
\centering
  \caption{A summary of several IML methods based on different choices of loss functions, regularizations and constraints under the JELLA framework.}\label{tab_summary}
\begin{tabular}{|c|c|c|c|c|c|c|}
  \hline
 & Methods & $f^{(v)}(\mathbf{Z}^{(v)}, \mathbf{U}^{(v)}\mathbf{W})$ & $\mathcal{R}_1$ & $\mathcal{R}_2$ & $\mathcal{C}_1^{(v)}$ & $\mathcal{C}_2$ \\ \hline
  \multirow{2}*{Originally  } & PVC\cite{Li2014PVC} & $\|\bar{\mathbf{X}}^{(v)} - \mathbf{U}^{(v)}\bar{\mathbf{W}}^{(v)}\|^2$ & - & $ \|\mathbf{W}\|_{1,1}$ & $\mathbf{U}^{(v)} \geq 0$ & $\mathbf{W} \geq 0$ \\ \cline{2-7}
 & MVL-IV\cite{xu2015MVL} & $\|\mathbf{Z}^{(v)} - \mathbf{U}^{(v)}\mathbf{W}\|^2$ & - & - & - & - \\ \cline{2-7}
 \multirow{3}*{designed for IML} & IMG\cite{Zhao2016IMG} & $\|\bar{\mathbf{X}}^{(v)} - \mathbf{U}^{(v)}\bar{\mathbf{W}}^{(v)}\|^2$ & $\|\mathbf{U}^{(v)}\|^2$ & LGR & - & - \\ \cline{2-7}
 & DAIMC \cite{Hu2018DAIMC} & $\|\bar{\mathbf{X}}^{(v)} - \mathbf{U}^{(v)}\bar{\mathbf{W}}^{(v)}\|^2$ & $\ell_{2,1}$-RR & - & - & $\mathbf{W} \geq 0$ \\ \cline{2-7}
 &  IML-BDR & $\|\mathbf{Z}^{(v)} - \mathbf{U}^{(v)}\mathbf{W}\|^2$ & -& BDR & - & - \\ \hline \hline
 \multirow{2}*{Adapted from complete} & \multirow{2}*{iRMKMC} & \multirow{2}*{ $(\alpha^{(v)})^{\gamma}\!\!\!\sum\limits_{i\in \Omega^{(v)}}\|\mathbf{x}_i^{(v)} - \mathbf{U}^{(v)}\mathbf{w}_i\|$} & \multirow{2}*{-}&\multirow{2}*{ -} &\multirow{2}*{ -} & $\mathbf{W} \in \{0, 1\}^{k\times n}$, \\
  & & & & & & $\mathbf{W}^T\mathbf{1} = \mathbf{1}$ \\ \cline{2-7}
  \multirow{2}*{multi-view methods}& \multirow{2}*{iMCL} & \multirow{2}*{$\|\mathbf{Z}^{(v)} - \mathbf{U}^{(v)}\mathbf{W}\|^2$} &  \multirow{2}*{$\|(\mathbf{U}^{(v)})^T\|_{\infty,1}$} & {$ \|\mathbf{W}\|_{1,1}$,} &  \multirow{2}*{$\mathbf{U}^{(v)} \geq 0$} & \multirow{2}*{$\mathbf{W} \in [0, 1]^{k\times n}$} \\
  & & & & LGR on labeled part & &\\ \hline

\end{tabular}
\end{table*}

\subsection{Unifying Existing IML Methods}\label{subsec_obs}
In this subsection, we analyze the relationship between the framework and some popular IML methods.
Concretely, Eq. (\ref{eq_general1}) or Eq. (\ref{eq_general2}) includes MVL-IV \cite{xu2015MVL}, PVC \cite{Li2014PVC}, IMG \cite{Zhao2016IMG} and DAIMC \cite{Hu2018DAIMC} as special cases.

Let $f^{(v)}(\mathbf{Z}^{(v)},\mathbf{U}^{(v)}\mathbf{W}) = \|\mathbf{Z}^{(v)} - \mathbf{U}^{(v)}\mathbf{W}\|^2$, it can be easily seen that MVL-IV (as shown in Eq. (\ref{eq_MVL})) \cite{xu2015MVL} is a special case of JELLA in Eq. (\ref{eq_general1}) with squared $\ell_2$-norm loss and without any constraint and regularization.

In the next, we formulate PVC, IMG and DAIMC under the general model.
As a result, PVC and IMG are naturally extended to the case with more than two views, and all the three methods can deal with the incomplete-view setting.

PVC is originally designed for two-view data in the missing-view setting.
Its formulation is shown in Eq. (\ref{eq_PVC}).
Note that the loss term of PVC is actually equal to $\sum\limits_{v=1}^V \sum\limits_{i \in \Omega^{(v)}}  \left\|\mathbf{x}_i^{(v)}- \mathbf{U}^{(v)}\mathbf{w}_i\right\|^2$ ($=\sum\limits_{v=1}^V \left\|\bar{\mathbf{X}}^{(v)}- \mathbf{U}^{(v)}\bar{\mathbf{W}}^{(v)}\right\|^2$) with $V = 2$.
Thus, Eq. (\ref{eq_PVC}) can be naturally extended to the multi-view case with the following compact form:
\begin{equation}\label{eq_multiPVC}
\begin{array}{rl}
\mathop {\min}\limits_{\{\mathbf{U}^{(v)}\}_{v=1}^V, \mathbf{W}} & \sum\limits_{v=1}^V \sum\limits_{i \in \Omega^{(v)}}  \|\mathbf{x}_i^{(v)}- \mathbf{U}^{(v)}\mathbf{w}_i\|^2 + \alpha \|\mathbf{W}\|_{1,1} \\
\text{s.t.}
& \mathbf{U}^{(v)} \geq 0, \mathbf{W} \geq 0, v = 1, \cdots, V, \\
\end{array}
\end{equation}
where $\mathbf{W} = [\mathbf{w}_1, \cdots, \mathbf{w}_n]$ is the latent representation.
Apparently, Eq. (\ref{eq_multiPVC}) is a special case of JELLA (Eq. (\ref{eq_general2})) in the missing-view setting.

Moreover, if the low-rank matrices $\{\mathbf{Z}^{(v)}\}_{v=1}^V$ are introduced to approximate the original data matrices, then the formulation of PVC for the incomplete-view setting is
\begin{equation}\label{eq_multiPVC2}
\begin{array}{l}
\mathop {\min}\limits_{\substack{\{\mathbf{Z}^{(v)}\}_{v=1}^V, \mathbf{W}\\\{\mathbf{U}^{(v)}\}_{v=1}^V}}  \sum\limits_{v=1}^V \left\|\mathbf{Z}^{(v)}- \mathbf{U}^{(v)}\mathbf{W}\right\|^2 + \alpha \|\mathbf{W}\|_{1,1} \\
\quad \text{s.t.} ~~~ \boldsymbol{\Gamma} \odot \mathbf{Z}^{(v)} = \boldsymbol{\Gamma} \odot \mathbf{X}^{(v)}, \mathbf{U}^{(v)} \geq 0, \mathbf{W} \geq 0, \forall v. \\
\end{array}
\end{equation}

IMG (Eq. (\ref{eq_IMG})) \cite{Zhao2016IMG} can be deemed to be an enhanced version of PVC.
It can be extended to the multi-view case in the same way with PVC. The extended formulations of IMG in the missing-view setting and the incomplete-view setting are
\begin{equation}\label{eq_multiIMG}
\begin{array}{l}
\mathop {\min}\limits_{\substack{\{\mathbf{U}^{(v)}\}_{v=1}^V, \mathbf{W},\\\mathbf{A}\mathbf{1} = \mathbf{1}, \mathbf{A} \geq 0}}  \sum\limits_{v=1}^V \sum\limits_{i \in \Omega^{(v)}}  \left\|\mathbf{x}_i^{(v)}- \mathbf{U}^{(v)}\mathbf{w}_i\right\|^2 \\
\qquad \qquad + \alpha \|\mathbf{U}^{(v)}\|^2 + \beta Tr(\mathbf{W}\mathbf{L}_{\mathbf{A}}\mathbf{W}^T) + \gamma\|\mathbf{A}\|^2, \\
\end{array}
\end{equation}
and
\begin{equation}\label{eq_multiIMG2}
\begin{array}{l}
\mathop {\min}\limits_{\substack{\{\mathbf{Z}^{(v)}\}_{v=1}^V, \mathbf{W}\\\{\mathbf{U}^{(v)}\}_{v=1}^V}}  \sum\limits_{v=1}^V \left\|\mathbf{Z}^{(v)}- \mathbf{U}^{(v)}\mathbf{W}\right\|^2 + \alpha \|\mathbf{U}^{(v)}\|^2\\
\qquad\qquad + \beta tr(\mathbf{W}\mathbf{L}_{\mathbf{A}}\mathbf{W}^T) + \gamma\|\mathbf{A}\|^2 \\
\qquad \text{s.t.} \qquad \boldsymbol{\Gamma} \odot \mathbf{Z}^{(v)} = \boldsymbol{\Gamma} \odot \mathbf{X}^{(v)}, \mathbf{A}\mathbf{1} = \mathbf{1}, \mathbf{A} \geq 0.\\
\end{array}
\end{equation}

With the same trick,  DAIMC (Eq. (\ref{eq_DAIMC})) \cite{Hu2018DAIMC} can also be unified into the JELLA framework for the missing-view setting, and extended to the incomplete-view setting.
The formulations are
\begin{equation}\label{eq_DAIMC2}
\begin{array}{l}
\mathop {\min}\limits_{\substack{\{\mathbf{U}^{(v)},\mathbf{B}^{(v)} \}_{v=1}^V, \\ \mathbf{W} \geq 0}}  \sum\limits_{v=1}^V \big\{\sum\limits_{i\in \Omega^{(v)}} \|\mathbf{x}_i^{(v)} - \mathbf{U}^{(v)}\mathbf{w}_i\|^2 \\
\qquad \qquad \qquad + \alpha (\|(\mathbf{B}^{(v)})^T\mathbf{U}^{(v)} - \mathbf{I}\|^2 + \beta \|\mathbf{B}^{(v)}\|_{2,1})\big\},\\
\end{array}
\end{equation}
and
\begin{equation}\label{eq_DAIMC3}
\begin{array}{l}
\mathop {\min}\limits_{\substack{\{\mathbf{U}^{(v)},\mathbf{B}^{(v)} \}_{v=1}^V, \\ \mathbf{W} \geq 0}}  \sum\limits_{v=1}^V \big\{\|\mathbf{Z}^{(v)} - \mathbf{U}^{(v)}\mathbf{W}\|^2 \\
\qquad \qquad \qquad + \alpha (\|(\mathbf{B}^{(v)})^T\mathbf{U}^{(v)} - \mathbf{I}\|^2 + \beta \|\mathbf{B}^{(v)}\|_{2,1})\big\}\\
\qquad \text{s.t.} \qquad \boldsymbol{\Gamma} \odot \mathbf{Z}^{(v)} = \boldsymbol{\Gamma} \odot \mathbf{X}^{(v)}.
\end{array}
\end{equation}

The extended PVC, IMG and DAIMC can be addressed using the optimization procedure in Algorithm \ref{alg1} or Algorithm \ref{alg2}.
For instance, we provide the updating steps for solving PVC in Eq. (\ref{eq_multiPVC}):
\begin{equation}\label{eq_multiPVC_updateU}
\mathbf{U}_{ij}^{(v)}\! \leftarrow\! \max\!\Big(\!0, \mathbf{U}_{ij}^{(v)} \!+\! \frac{\big(\!\mathbf{U}^{(v)}\bar{\mathbf{W}}^{(v)}(\!\bar{\mathbf{W}}^{(v)}\!)^T \!-\! \bar{\mathbf{X}}^{(v)}(\!\bar{\mathbf{W}}^{(v)}\!)^T\!\big)_{ij}}{(\!\bar{\mathbf{W}}^{(v)}(\!\bar{\mathbf{W}}^{(v)}\!)^T\!)_{jj}}\!\Big),
\end{equation}
\begin{equation}\label{eq_multiPVC_updateP}
\mathbf{w}_{ij} \leftarrow \max\Big(0, \mathbf{w}_{ij} + \frac{\mathbf{b}_{ij} + 2\alpha}{\mathbf{C}_{jj}}\Big),
\end{equation}
where $\mathbf{b}_i =  \sum\limits_{v\in\{v'|i\in \Omega^{(v')}\}} \big(\mathbf{x}_i^{(v)}\mathbf{U}^{(v)} - \mathbf{U}^{(v)}(\mathbf{U}^{(v)})^T\mathbf{w}_i\big)$, $\mathbf{C} = \sum\limits_{v\in\{v'|i\in \Omega^{(v')}\}}\mathbf{U}^{(v)}(\mathbf{U}^{(v)})^T$, and $\{v'|i \in \Omega^{(v')}\}$ denotes the set of views where $\mathbf{x}_i$ is not missing.

\subsection{Adapting complete multi-view methods to IML}\label{subsec_pRMKMC}
In this subsection, we show that with the similar spirit of the framework, some existing complete multi-view approaches can be adapted to IML.
In particular, RMKMC \cite{Cai2013MvKmeans} and MCL \cite{Guan2015MCL}, which learn a unified pattern from multiple views by linear transformation, are adapted for demonstration.

\paragraph{RMKMC for the missing-view setting.}
To make the algorithm more robust to outliers, RMKMC \cite{Cai2013MvKmeans} utilizes the structured sparsity-inducing norm to combine multiple views together.
The formulation of RMKMC is
\begin{equation}\label{eq_RMKMC}
\begin{array}{l}
\mathop {\min} \limits_{\mathbf{U}^{(v)}, \mathbf{W}, \alpha^{(v)}}  \sum\limits_{v=1}^V (\alpha^{(v)})^{\gamma}\|\mathbf{X}^{(v)} -\mathbf{U}^{(v)}\mathbf{W} \|_{2,1}\\
 \text{s.t.}~ \mathbf{W} \in \{0, 1\}^{k\times n}, \mathbf{W}^T\mathbf{1} = \mathbf{1}, \sum\nolimits_{v=1}^V\alpha^{(v)} = 1,\\
\end{array}
\end{equation}
where $\mathbf{U}^{(v)}$ is the centroid matrix for the $v$-th view, $\mathbf{W} = [\mathbf{w}_1, \cdots, \mathbf{w}_n]$ is the cluster indicator matrix (i.e., $\mathbf{W}_{ji} = 1$ if the $i$-th point belongs to the $j$-th cluster, and 0 otherwise), $k$ is the number of clusters, $\alpha^{(v)} \geq 0 $ is the weight for the $v$-th view, and $\gamma \geq 1$ is the parameter to control the weight distribution.

Respectively, taking the cluster indicator matrix and the centroid matrices as the latent embedding and the transformation matrices in Eq. (\ref{eq_general2}), the objective of RMKMC for the missing-view setting\footnote{Since $\mathbf{W}$ is constrained to be the cluster indicator matrix, the recovery capability of the model may be limited if we introduce the low-rank approximate matrices. Hence, here we only adapt RMKMC to the missing-view setting.} can be written as
\begin{equation}\label{eq_pRMKMC}
\begin{array}{l}
\mathop {\min} \limits_{\mathbf{U}^{(v)}, \mathbf{W}, \alpha^{(v)}}  \sum\limits_{v=1}^V\sum\limits_{i\in \Omega^{(v)}} (\alpha^{(v)})^{\gamma}\|\mathbf{x}_i^{(v)} -\mathbf{U}^{(v)}\mathbf{w}_i \|\\
\text{s.t.} ~ \mathbf{W} \in \{0, 1\}^{k\times n}, \mathbf{W}^T\mathbf{1} = \mathbf{1}, \sum\nolimits_{v=1}^V\alpha^{(v)} = 1,\\
\end{array}
\end{equation}
which is referred to as incomplete RMKMC (iRMKMC) for convenience.

\paragraph{Incomplete MCL.}
Multi-view Concept Learning (MCL) \cite{Guan2015MCL} is a semi-supervised nonnegative latent representation learning algorithm for multi-view data.
To preserve the semantic relationships between labeled samples and explore the association between latent components and views, MCL imposes the graph regularization on the labeled samples' representation matrix and adds structured sparsity constraints on the basis matrices.
Specifically, its formulation is
\begin{equation}\label{MCL}
\begin{array}{l}
\mathop {\min }\limits_{\scriptstyle\{ {\mathbf{U}^{(v)}}\},\hfill\atop
\scriptstyle \quad \mathbf{W},\boldsymbol{\alpha} \hfill}  \frac{1}{2}\sum\limits_{v = 1}^V \| \mathbf{X}^{(v)} - \mathbf{U}^{(v)}\mathbf{W}  \|^2 + \alpha \sum\limits_{v = 1}^V\|(\mathbf{U}^{(v)})^T\|_{\infty,1}\\
\qquad \qquad + \frac{\beta}{2}Tr(\mathbf{W}_l(\mathbf{L}_w - \mathbf{L}_b)\mathbf{W}_l^T) + \gamma\|\mathbf{W}\|_{1,1}\\
\text{s.t.}  \quad \mathbf{U}^{(v)}\geq 0, 1\geq\mathbf{W}_{ij} \geq 0, \forall i,j,v,
\end{array}
\end{equation}
where $\mathbf{W}_l$ denotes the embedding of the labeled points, $\mathbf{L}_w$ and $\mathbf{L}_b$ are the Laplacian matrices for the within-class affinity graph and between-class penalty graph, respectively, $\alpha$, $\beta$ and $\gamma$ are positive parameters.

To adapt MCL to IML, we just need introducing the low-rank matrices $\{\mathbf{Z}^{(v)}\}_{v=1}^V$ and the corresponding constraints.
Thus, the formulation of the incomplete MCL (iMCL) is
\begin{equation}\label{iMCL}
\begin{array}{l}
\mathop {\min }\limits_{\scriptstyle\{ {\mathbf{U}^{(v)}}\},\hfill\atop
\scriptstyle \quad \mathbf{W},\boldsymbol{\alpha} \hfill}  \frac{1}{2}\sum\limits_{v = 1}^V \| \mathbf{Z}^{(v)} - \mathbf{U}^{(v)}\mathbf{W}  \|^2 + \alpha \sum\limits_{v = 1}^V\|(\mathbf{U}^{(v)})^T\|_{\infty,1}\\
\qquad \qquad + \frac{\beta}{2}Tr(\mathbf{W}_l(\mathbf{L}_w - \mathbf{L}_b)\mathbf{W}_l^T) + \gamma\|\mathbf{W}\|_{1,1}\\
\text{s.t.}  ~~ \boldsymbol{\Gamma} \odot \mathbf{Z}^{(v)}\! =\! \boldsymbol{\Gamma} \odot \mathbf{X}^{(v)}, \mathbf{U}^{(v)}\geq 0, 1\!\geq\mathbf{W}_{ij} \geq 0, \forall i,j,v.
\end{array}
\end{equation}

In summary, the analysis in Sec. \ref{subsec_obs} and Sec. \ref{subsec_pRMKMC} indicates that JELLA is a unified framework in viewing different IML methods, which are originally designed for IML or adapted from complete multi-view methods.
Table \ref{tab_summary} presents a summary of these special cases of JELLA with different loss functions, regularizations and constraints.

It also can be seen that one can cope with incomplete multi-view data immediately, by adapting complete multi-view methods to IML within the JELLA framework.
Compared with designing new algorithms, the efficiency of dealing with incomplete multi-view data is largely improved.
Therefore, the proposed JELLA framework is of practical significance.
\section{Incomplete Multi-view Learning with Block Diagonal Representation}\label{sec_IML_BDR}
In this section, within the JELLA framework, we formulate a specific model with the Block Diagonal Representation (BDR) for IML.

\subsection{The Method}
In the proposed method, we assume that the incomplete multi-view data are generated from a union of $k$ subspaces.
Correspondingly, the learned unified and full embedding $\mathbf{W}$ is seen as the authentic samples lying exactly on the subspaces.
A recently research reveals that method with the block diagonal property would possibly lead to correct subspace clustering \cite{Lu2018BDR}.
To increase the discriminability of the learned embedding $\mathbf{W}$, we introduce the $k$-block diagonal representation matrix \cite{Lu2018BDR} $\mathbf{B} \in \mathbb{R}^{n \times n}$ to self-express $\mathbf{W}$, i.e., $\mathbf{W} = \mathbf{W}\mathbf{B}$.
To ensure the $k$-block diagonal property of $\mathbf{B}$, the $k$-block diagonal regularizer is exploited.

\begin{define}[$k$-block diagonal regularizer, \cite{Lu2018BDR}] \label{def_BDR}
Given a similarity matrix $\mathbf{B} \in \mathbb{R}^{n \times n}$, the $k$-block diagonal regularizer is defined as the sum of the $k$ smallest eigenvalues of $\mathbf{L}_\mathbf{B}$, i.e.,
\begin{equation}\label{eq_BDR}
\|\mathbf{B}\|_{ {\fbox{k}}} = \sum\limits_{i=1}^k \sigma_i(\mathbf{L}_\mathbf{B}),
\end{equation}
where $\mathbf{L}_\mathbf{B} = \text{Diag}(\mathbf{B}\mathbf{1}) -\mathbf{B}$ is the Laplacian matrix of $\mathbf{B}$, and $\sigma_i(\mathbf{L}_\mathbf{B})$ is the $i$-th smallest eigenvalue of $\mathbf{L}_\mathbf{B}$.
\end{define}

Substituting the Block Diagonal Representation (BDR) term (i.e., self-expression term and the block diagonal regularizer) into the JELLA framework, we obtain the formulation of the Incomplete Multi-view Learning with Block Diagonal Representation (IML-BDR) algorithm:
\begin{equation}\label{IML_BDR}
\begin{array}{l}
\mathop {\min}\limits_{\substack{\mathbf{W},\mathbf{B}\\\{\mathbf{U}^{(v)}, \mathbf{Z}^{(v)}\}  }} \!\! \sum\limits_{v = 1}^V \left\|\mathbf{Z}^{(v)} - \mathbf{U}^{(v)}\mathbf{W}\right\|^2 \!+\! \alpha \|\mathbf{W}-\mathbf{W}\mathbf{B}\|^2
 \!+\!\gamma\|\mathbf{B}\|_{\fbox{\textit{k}}}\\
\quad \text{s.t.}\quad \mathbf{B}\in \mathcal{B}, \boldsymbol{\Gamma}^{(v)} \odot \mathbf{Z}^{(v)} = \boldsymbol{\Gamma}^{(v)} \odot \mathbf{X}^{(v)} , \forall v ,\\
\end{array}
\end{equation}
where $\mathcal{B}=\{\mathbf{B}|\text{diag}(\mathbf{B}) = 0, \mathbf{B} = \mathbf{B}^T, \mathbf{B}\geq 0\}$, $\alpha$ and $\gamma$ are positive parameters.
To maintain the reconstruction performance of IML-BDR, no constraint is imposed on the transformation matrix $\mathbf{U}^{(v)}$.
Since $\mathbf{B}$ is an affinity matrix, it is required to be nonnegative and symmetric.
Noting that these constraints on $\mathbf{B}$ will restrict its capability in representation, an intermediate term $\mathbf{P}$ is introduced
\begin{equation}\label{IML_BDR2}
\begin{array}{l}
\mathop {\min}\limits_{\substack{ \mathbf{W},\mathbf{P} ,\mathbf{B}\\\{\mathbf{U}^{(v)}, \mathbf{Z}^{(v)}\}}}  \!\! \sum\limits_{v = 1}^V \left\|\mathbf{Z}^{(v)} - \mathbf{U}^{(v)}\mathbf{W}\right\|^2 \!+\! \alpha \|\mathbf{W}-\mathbf{W}\mathbf{P}\|^2    \\
\qquad \qquad +\beta\|\mathbf{P} - \mathbf{B}\|^2+\gamma\|\mathbf{B}\|_{\fbox{\textit{k}}}\\
\text{s.t.}\quad \mathbf{B}\in \mathcal{B}, \boldsymbol{\Gamma}^{(v)} \odot \mathbf{Z}^{(v)} = \boldsymbol{\Gamma}^{(v)} \odot \mathbf{X}^{(v)} ,\forall v.\\
\end{array}
\end{equation}
Problem (\ref{IML_BDR2}) is equivalent to problem  (\ref{IML_BDR}) when $\beta > 0$ is sufficiently large.
Moreover, as we will show in next subsection, the term $\beta\|\mathbf{P} - \mathbf{B}\|^2$ makes the subproblems with respect to (w.r.t.) $\mathbf{P}$ and $\mathbf{B}$ strongly convex.

\subsection{Solution to IML-BDR}\label{sec_opt}
The main difficulty to address problem (\ref{IML_BDR2}) is due to the non-convex term $\|\mathbf{B}\|_{\fbox{\textit{k}}}$.
According to the Ky Fan's Theorem \cite{Fan1949}, we have
\begin{equation}
\|\mathbf{B}\|_{\fbox{\textit{k}}} = \sum\limits_{i=1}^k \sigma_i(\mathbf{L}_\mathbf{B}) = \min\limits_{\mathbf{F}\in \mathcal{F}} Tr(\mathbf{F}^T\mathbf{L}_{\mathbf{B}}\mathbf{F}),
\end{equation}
where $\mathcal{F} = \{\mathbf{F}|\mathbf{F}\in \mathbb{R}^{n\times k}, \mathbf{F}^T\mathbf{F} = \mathbf{I}\}$.
Therefore, the problem (\ref{IML_BDR2}) is equivalent to
\begin{equation}\label{IML_BDR3}
\begin{array}{l}
\mathop {\min}\limits_{\substack{\mathbf{W},\mathbf{P}, \mathbf{B}, \mathbf{F},\\\{\mathbf{U}^{(v)}, \mathbf{Z}^{(v)}\}}}\!\!  \sum\limits_{v = 1}^V \! \left\|\mathbf{Z}^{(v)} \! -\!  \mathbf{U}^{(v)}\mathbf{W}\right\|^2 \! +\!  \alpha \|\mathbf{W}\! -\! \mathbf{W}\mathbf{P}\|^2    \\
\qquad\qquad +\beta\|\mathbf{P} - \mathbf{B}\|_F^2+\gamma Tr(\mathbf{F}^T\mathbf{L}_{\mathbf{B}}\mathbf{F})\\
\text{s.t.} ~ \mathbf{B}\in \mathcal{B}, \mathbf{F}\in \mathcal{F},  \boldsymbol{\Gamma}^{(v)} \odot \mathbf{Z}^{(v)} = \boldsymbol{\Gamma}^{(v)} \odot \mathbf{X}^{(v)} , \forall v.
\end{array}
\end{equation}

Following Algorithm \ref{alg1}, the alternating minimization strategy is adopted to address Eq. (\ref{IML_BDR3}).
With the current solutions $\{\mathbf{U}_t^{(v)}, \mathbf{Z}_t^{(v)}\}_{v=1}^V$, $\mathbf{W}_t$, $\mathbf{P}_t$, $\mathbf{B}_t$ and $\mathbf{F}_t$, we update each variable separately by minimizing Eq. (\ref{IML_BDR3}) with the other variables being fixed as constant.
Concretely, the solutions to each variable is obtained by addressing the following subproblems in sequence:
\begin{align}
&\mathbf{U}_{t+1}^{(v)} \!=  \mathop {\arg\min}\limits_{\mathbf{U}^{(v)} \in \mathbb{R}^{d^{(v)}\times r}} \|\mathbf{Z}_t^{(v)} \!-\! \mathbf{U}^{(v)}\mathbf{W}_t\|^2 \label{eq_sub1}\\
& \qquad    =  \mathbf{Z}_t^{(v)}\mathbf{W}_t^T(\mathbf{W}_t\mathbf{W}_t^T)^{\dagger}, \forall v,\notag \\
&\mathbf{W}_{t+1} \!=\!\mathop {\arg\min}\limits_{\mathbf{W}\in \mathbb{R}^{ r \times n}} \!\sum\limits_{v = 1}^V \|\mathbf{Z}_t^{(v)} \!-\! \mathbf{U}_{t+1}^{(v)}\mathbf{W}\|^2 \! +\! \alpha \|\mathbf{W}\!-\!\mathbf{W}\mathbf{P}_t\|^2, \label{eq_sub2}\\
& \mathbf{P}_{t+1} =\mathop {\arg\min}\limits_{\mathbf{P}\in \mathbb{R}^{ n \times n}}  \|\mathbf{W}_{t+1}\!-\!\mathbf{W}_{t+1}\mathbf{P}\|^2 \!+\! \frac{\beta}{\alpha} \|\mathbf{P} \!-\! \mathbf{B}_t\|^2, \label{eq_sub3}\\
&\qquad  = (\mathbf{W}_{t+1}^T\mathbf{W}_{t+1} + \frac{\beta}{\alpha} \mathbf{I})^{-1}(\mathbf{W}_{t+1}^T\mathbf{W}_{t+1} + \frac{\beta}{\alpha} \mathbf{B}_t), \notag \\
& \mathbf{B}_{t+1} =\mathop {\arg\min}\limits_{\mathbf{B}\in \mathcal{B}}   \beta \|\mathbf{P}_{t+1} - \mathbf{B}\|^2 +\gamma Tr(\mathbf{F}_t^T\mathbf{L}_{\mathbf{B}}\mathbf{F}_t), \label{eq_sub4}\\
& \mathbf{F}_{t+1} =\mathop {\arg\min}\limits_{\mathbf{F}\in \mathcal{F}} Tr(\mathbf{F}^T\mathbf{L}_{\mathbf{B}_{t+1}}\mathbf{F}), \label{eq_sub5} \\
& \mathbf{Z}_{t+1}^{(v)} = \mathop {\arg\min}\limits_{\boldsymbol \Gamma^{(v)}\odot(\mathbf{Z}^{(v)} - \mathbf{X}^{(v)}) = \mathbf{0}} \|\mathbf{Z}^{(v)} - \mathbf{U}_{t+1}^{(v)}\mathbf{W}_{t+1}\|^2, \forall v.\label{eq_sub6}
\end{align}

Setting the derivative of Eq. (\ref{eq_sub2}) w.r.t. $\mathbf{W}$ to zeros, we have
\begin{equation}\label{eq_V}
\big(\sum\limits_{v=1}^V (\mathbf{U}_{t+1}^{(v)})^T\mathbf{U}_{t+1}^{(v)}\big) \mathbf{W} + \alpha\mathbf{W}(\mathbf{I} - \mathbf{P}_t)^2  = \sum\limits_{v=1}^V (\mathbf{U}_{t+1}^{(v)})^T\mathbf{Z}_{t}^{(v)}.
\end{equation}
Eq. (\ref{eq_V}) is a Sylvester equation, and its solution is unique while the spectra of $\sum\limits_{v=1}^V (\mathbf{U}_{t+1}^{(v)})^T\mathbf{U}_{t+1}^{(v)}$ and $-\alpha(\mathbf{I} - \mathbf{P}_t)^2$ are nonoverlapping \cite{Bartels1972SylvesterEq}.
For convenience, we use $Syl( \mathbf{W}; \mathbf{U}_{t+1}^{(v)}, \mathbf{Z}_{t}^{(v)}, \mathbf{P}_{t})$ to denote the Sylvester equation defined in Eq. (\ref{eq_V}).

The B-subproblem in Eq. (\ref{eq_sub4}) is equivalent to
\[
    \mathbf{B}_{t+1} =\mathop {\arg\min}\limits_{\mathbf{B}\in \mathcal{B}}   \|\mathbf{B} - \mathbf{P}_{t+1} +\frac{\gamma}{2\beta}(\text{diag}(\mathbf{F}_t^T\mathbf{F}_t)\mathbf{1}^T - \mathbf{F}_t^T\mathbf{F}_t) \|^2,
\]
whose solution can be obtained in a closed form by using the following lemma \cite{Lu2018BDR}.
\begin{lemma}[\cite{Lu2018BDR}]
Given $\mathbf{A}\in \mathbb{R}^{ n \times n}$. Let $\hat{\mathbf{A}} = \mathbf{A} - \text{Diag}(\text{diag}(\mathbf{A}))$, then the solution to
$
\min\limits_{\mathbf{B}\in \mathcal{B}} \|\mathbf{B} - \mathbf{A}\|^2
$
is given by $\mathbf{B}^* = \left[(\hat{\mathbf{A}} + \hat{\mathbf{A}} ^T)/2\right]_{+}$.
\end{lemma}
Denote $\mathbf{Q} = \mathbf{P}_{t+1} - \frac{\gamma}{2\beta}\left(\text{diag}(\mathbf{F}_t^T\mathbf{F}_t)\mathbf{1}^T - \mathbf{F}_t^T\mathbf{F}_t\right)$ and $\hat{\mathbf{Q}} = \mathbf{Q} - \text{Diag}(\text{diag}(\mathbf{Q}))$.
$\mathbf{B}_{t+1}$ is updated by $\mathbf{B}_{t+1} = \left[(\hat{\mathbf{Q}} + \hat{\mathbf{Q}} ^T)/2\right]_{+}$.

Then, the optimal solution $\mathbf{F}_{t+1}$ is formed by the $k$ eigenvectors corresponding to the $k$ smallest eigenvalues of $\mathbf{L}_{\mathbf{B}_{t+1}}$.
Finally, $\mathbf{Z}_{t+1}^{(v)}$ is updated by $\mathbf{Z}_{t+1}^{(v)} = \mathbf{U}_{t+1}^{(v)}\mathbf{W}_{t+1} + \boldsymbol{\Gamma}^{(v)} \odot (\mathbf{X}^{(v)} - \mathbf{U}_{t+1}^{(v)}\mathbf{W}_{t+1})$ ($\forall v$).

\subsection{An Accelerated Implementation}
The basic procedures in Eqs. (\ref{eq_sub1}) - (\ref{eq_sub6}) is commonly used and reliable, but it is of low efficiency for large data.
To speed up the convergence, we incorporate the successive over-relaxation (SOR) \cite{xu2015MVL,Wen2012SOR} method into the basic procedures in Eqs. (\ref{eq_sub1}) - (\ref{eq_sub6}).
The SOR method is generalized from the Gauss-Seidel method by using the extrapolation method \cite{Cullum1996GS}.
When searching for $\mathbf{U}_{t+1}^{(v)}$ from $\mathbf{U}_{t}^{(v)}$, a certain amount is moved along the direction $\mathbf{U}_{t+1}^{(v)} - \mathbf{U}_{t}^{(v)}$.
Note that $\mathbf{U}_{t+1}^{(v)} = \mathbf{U}_{t}^{(v)} + (\mathbf{U}_{t+1}^{(v)} - \mathbf{U}_{t}^{(v)})$.
Assume that the direction $\mathbf{U}_{t+1}^{(v)} - \mathbf{U}_{t}^{(v)}$ takes us closer, but not always, to the truth. Then, there might be advantages by moving along the direction $\mathbf{U}_{t+1}^{(v)} - \mathbf{U}_{t}^{(v)}$ more far away, i.e., $\mathbf{U}_{t+1}^{(v)} = \mathbf{U}_{t}^{(v)} + \lambda (\mathbf{U}_{t+1}^{(v)} - \mathbf{U}_{t}^{(v)})$ with $\lambda > 1$.
This iterative step reduces to the Gauss-Seidel method when $\lambda = 1$.
It has been suggested that the convergence from $\mathbf{U}_{t+1}^{(v)}$ to its ground truth is usually faster if we use the SOR-like updating scheme.

By exploiting the SOR technique, the new updating equation for $\mathbf{U}^{(v)}$ is
\begin{equation}\label{eq_Ul}
\mathbf{U}_{t+1,\lambda}^{(v)} = \lambda\mathbf{U}_{t+1}^{(v)} + (1-\lambda)\mathbf{U}_{t}^{(v)},  \lambda \geq 1.
\end{equation}
Define the residual on the $v$-th view as $ \mathbf{R}^{(v)} = \boldsymbol{\Gamma}^{(v)}\odot(\mathbf{X}^{(v)} - \mathbf{U}^{(v)}\mathbf{W})$.
According to Eq. (\ref{eq_sub6}), we have $\mathbf{Z}_{t+1}^{(v)} = \mathbf{U}_{t+1,\lambda}^{(v)}\mathbf{W}_{t+1} + \mathbf{R}_{t+1}^{(v)}$.
Define
\begin{equation}\label{eq_Zl1}
\mathbf{Z}_{t+1,\lambda}^{(v)} = \mathbf{U}_{t+1,\lambda}^{(v)}\mathbf{W}_{t+1} + \lambda \mathbf{R}_{t+1}^{(v)}.
\end{equation}
Since
\begin{equation}\label{eq_Zl2}
\lambda\mathbf{Z}_{t+1}^{(v)} = \lambda\mathbf{U}_{t+1,\lambda}^{(v)}\mathbf{W}_{t+1} + \lambda\mathbf{R}_{t+1}^{(v)},
\end{equation}
then, we have
\begin{equation}\label{eq_Zl3}
\mathbf{Z}_{t+1,\lambda}^{(v)}  = \lambda\mathbf{Z}_{t+1}^{(v)} + (1 - \lambda)\mathbf{U}_{t+1,\lambda}^{(v)}\mathbf{W}_{t+1}.
\end{equation}
Multiplying both sides of Eq. (\ref{eq_Zl3}) with $\mathbf{W}_{t+1}^T(\mathbf{W}_{t+1}\mathbf{W}_{t+1}^T)^{\dagger}$, we obtain
\begin{equation}\label{eq_Zl4}
\begin{array}{l}
\mathbf{Z}_{t+1,\lambda}^{(v)}\!\mathbf{W}_{t+1}^T(\!\mathbf{W}_{t+1, \lambda}\!\mathbf{W}_{t+1}^T\!)^{{\dagger}} \\
= \lambda \mathbf{U}_{t+2}^{(v)} + (1 - \lambda)\mathbf{U}_{t+1,\lambda}^{(v)} = \mathbf{U}_{t+2,\lambda}^{(v)}.
\end{array}
\end{equation}
Thus, the steps in Eqs. (\ref{eq_sub1}) and (\ref{eq_Ul}) can be merged, and $\mathbf{U}_{t+1, \lambda}^{(v)}$ is updated by
\begin{equation}
\mathbf{U}_{t+1, \lambda}^{(v)} = \mathbf{Z}_{t, \lambda}^{(v)}\mathbf{W}_t^T(\mathbf{W}_t\mathbf{W}_t^T)^{\dagger}.
\end{equation}
$\mathbf{W}_{t+1}$, $\mathbf{P}_{t+1}$, $\mathbf{B}_{t+1}$, $\mathbf{F}_{t+1}$ and $\mathbf{Z}_{t+1}^{(v)}$ are updated by solving subproblems (\ref{eq_sub2}) - (\ref{eq_sub6}) as before.
Then we calculate $\mathbf{Z}_{t+1,\lambda}^{(v)}$ according to Eq. (\ref{eq_Zl1}). The whole procedure is as follows.
\begin{align}
& \mathbf{U}_{t+1, \lambda}^{(v)} = \mathbf{Z}_{t, \lambda}^{(v)}\mathbf{W}_t^T(\mathbf{W}_t\mathbf{W}_t^T)^{\dagger}\label{eq_sor1}\\
& \mathbf{W}_{t+1} = \text{solution to } Syl( \mathbf{W}; \mathbf{U}_{t+1,\lambda}^{(v)}, \mathbf{Z}_{t, \lambda}^{(v)}, \mathbf{P}_{t}), \label{eq_sor2}\\
& \mathbf{P}_{t+1} = (\mathbf{W}_{t+1}^T\mathbf{W}_{t+1} + \frac{\beta}{\alpha} \mathbf{I})^{-1}(\mathbf{W}_{t+1}^T\mathbf{W}_{t+1} + \frac{\beta}{\alpha} \mathbf{B}_t), \label{eq_sor3} \\
& \mathbf{B}_{t+1} = \left[(\hat{\mathbf{Q}} + \hat{\mathbf{Q}} ^T)/2\right]_{+}, \Big(\hat{\mathbf{Q}} = \mathbf{Q} -  \text{Diag}(\text{diag}(\mathbf{Q})) \label{eq_sor4}\\
& \qquad  \mathbf{Q} = \mathbf{P}_{t+1} - \frac{\gamma}{2\beta}\left(\text{diag}(\mathbf{F}_t^T\mathbf{F}_t)\mathbf{1}^T - \mathbf{F}_t^T\mathbf{F}_t\right)\Big), \notag \\
&\mathbf{F}_{t+1} =\mathop{ \arg\min} \nolimits_{\mathbf{F}\in\mathcal{F}} tr(\mathbf{F}^T\mathbf{L}_{\mathbf{B}_{t+1}}\mathbf{F}),\label{eq_sor5}\\
&\mathbf{Z}_{t+1}^{(v)} \!=\! \mathbf{U}_{t+1,\lambda}^{(v)}\mathbf{W}_{t+1} \!+ \! \boldsymbol{\Gamma}^{(v)}\! \odot \!(\mathbf{X}^{(v)}\! -\! \mathbf{U}_{t+1,\lambda}^{(v)}\mathbf{W}_{t+1}),\label{eq_sor6}\\
&\mathbf{Z}_{t+1,\lambda}^{(v)} =  \lambda\mathbf{Z}_{t+1}^{(v)} + (1 - \lambda)\mathbf{U}_{t+1,\lambda}^{(v)}\mathbf{W}_{t+1}\label{eq_sor7}.
\end{align}

$\lambda$ controls the amount that we exceed the standard Gauss-Seidel correction. It is usually not good enough to use a fixed $\lambda$.
Hence, we adjust $\lambda$ accordingly based on the change of two consecutive objective values.
More concretely, we calculate the ratio of two consecutive objective values after all variables are updated:
\begin{equation}\label{eq_resRate}
\rho(\lambda)
=  \frac{g_{t+1}}{g_t},
\end{equation}
where $g_t = g(\{\mathbf{Z}_{t}^{(v)}\}_{v=1}^V, \{\mathbf{U}_{t}^{(v)}\}_{v=1}^V, \mathbf{W}_{t}, \mathbf{P}_{t}, \mathbf{B}_{t}, \mathbf{F}_{t})$ denotes the objective value of the $t$-th iteration.

$\rho(\lambda) < 1$ means that the objective value is decreased and the currently obtained point is acceptable.
Otherwise, we just need to set $\lambda = 1$ and run the steps in Eqs. (\ref{eq_sub1}) - (\ref{eq_sub6}), then, $\rho(\lambda) < 1$ is guaranteed.
$\rho(\lambda)$ measures the degree of decrease in objective values brought by $\lambda$. If $\rho(\lambda)$ is small, then it suggests that the current $\lambda$ is effective and can be remained unchanged.
When $\rho(\lambda) < 1$ but is larger than a threshold $\rho_1$ ($0<\rho_1 < 1$), it is deemed that the objective value is not decreased enough.
Thus, $\lambda$ is increased to $\min(\lambda + \delta, \lambda_{max})$, where $\delta > 0$ is the step size and $\lambda_{max}$ is the allowed maximum value for $\lambda$.
Algorithm \ref{alg:IML_BDR} summarizes the above SOR-like optimization procedure for the IML-BDR approach.

\begin{algorithm}
\caption{SOR-like optimization for IML-BDR}
\label{alg:IML_BDR}
\begin{algorithmic}
\STATE \textbf{Input:}
$\{\mathbf{X}^{(v)}\}_{v=1}^V$, $\{\mathbf{\Gamma}^{(v)} \}_{v=1}^V$, initial $\{\mathbf{Z}_{0, \lambda}^{(v)} = \mathbf{Z}_0^{(v)} \}_{v=1}^V$ and $\mathbf{W}_0$, $\alpha > 0$ and $\beta > 0$, $\lambda = 1$, $\rho_1 = 0.7$, $\delta =0.2$, $\lambda_{max} = 5$, $t = 0$.
\STATE \textbf{Output:}
$\mathbf{W}$, $\mathbf{F}$, $\mathbf{U}^{(v)}$, and $\mathbf{Z}^{(v)} (\forall v \in [1, V])$.
\STATE \textbf{while} not meeting the stoping criterion \textbf{do}
\STATE 1: \quad \textbf{if} $t > 0$
\STATE 2: \qquad Compute $\mathbf{Z}_{t+1,\lambda}^{(v)}$ according to Eq. (\ref{eq_sor7}).
\STATE 3: \quad \textbf{end if}
\STATE 4: \quad Update $\mathbf{U}_{t+1,\lambda}^{(v)}$, $\forall v \in [1, V]$ by Eq. (\ref{eq_sor1}).
\STATE 5: \quad Update $\mathbf{W}_{t+1}$ by Eq. (\ref{eq_sor2}).
\STATE 6: \quad Update $\mathbf{P}_{t+1}$ by Eq. (\ref{eq_sor3}).
\STATE 7: \quad Update $\mathbf{B}_{t+1}$ by Eq. (\ref{eq_sor4}).
\STATE 8: \quad Update $\mathbf{F}_{t+1}$ by Eq. (\ref{eq_sor5}).
\STATE 9: \quad Update $\mathbf{Z}_{t+1}^{(v)}$, $\forall v \in [1, V]$ by Eq. (\ref{eq_sor6}).
\STATE 10: ~Compute $\rho(\lambda)$ according to Eq. (\ref{eq_resRate}).
\STATE 11: ~\textbf{if} $\rho(\lambda) \geq 1$
\STATE 12: \qquad Set $\lambda = 1$; \textbf{continue};
\STATE 13: ~\textbf{elseif} $\rho(\lambda) \geq \rho_1$
\STATE 14: \qquad $\lambda = \min(\lambda+\delta, \lambda_{max})$, $k = k + 1$.
\STATE 15: ~\textbf{else}
\STATE 16: \qquad $t = t+1$.
\STATE 17: ~\textbf{end if}
\STATE \textbf{end while}
\end{algorithmic}
\end{algorithm}

\subsection{Convergence Analysis}
Note that each iterative step of the basic algorithm in Eqs. (\ref{eq_sub1}) - (\ref{eq_sub6}) will obtain the global solution to the corresponding subproblems. Thus, the procedures in Eqs. (\ref{eq_sub1}) - (\ref{eq_sub6}) will not increase the objective value of Eq. (\ref{IML_BDR3}).
Now, we look at the SOR-like optimization in Algorithm \ref{alg:IML_BDR}.
After each iteration, the ratio $\rho(\lambda)$ of two consecutive objective values is calculated.
When this ratio is larger than 1, then the algorithm will go back to the basic algorithm in Eqs. (\ref{eq_sub1}) - (\ref{eq_sub6}) (Lines 11 - 12 of Algorithm \ref{alg:IML_BDR}).
Since the basic algorithm will not increase the objective value, it is easy to see that Algorithm \ref{alg:IML_BDR} will also not increase the objective value of Eq. (\ref{IML_BDR3}).
Note that $\mathbf{L}_\mathbf{B}$ is positive semi-definite.
It holds that $Tr(\mathbf{F}^T\mathbf{L}_{\mathbf{B}}\mathbf{F}) \geq \min\limits_{\mathbf{F}\in \mathcal{F}} Tr(\mathbf{F}^T\mathbf{L}_{\mathbf{B}}\mathbf{F}) = \sum\limits_{i=1}^k \sigma_i(\mathbf{L}_\mathbf{B})=\|\mathbf{B}\|_ {\fbox{\textit{k}}} \geq 0 $.
Thus, Eq. (\ref{IML_BDR3}) is lower bounded by 0.
We have the following conclusion.

\begin{prop}\label{obj_conv}
The sequence $\big\{\{\mathbf{Z}_{t}^{(v)}\}_{v=1}^V$, $\{\mathbf{U}_{t}^{(v)}\}_{v=1}^V$, $\mathbf{W}_{t}$, $\mathbf{P}_{t}, \mathbf{B}_{t}, \mathbf{F}_{t}\big\}$ generated by Algorithm \ref{alg:IML_BDR}  has the following properties:
\begin{enumerate}
\item The objective $g(\{\mathbf{Z}_{t}^{(v)}\}_{v=1}^V, \{\mathbf{U}_{t}^{(v)}\}_{v=1}^V, \mathbf{W}_{t}, \mathbf{P}_{t}, \mathbf{B}_{t}, \mathbf{F}_{t})$ is  monotonically decreasing, and the sequence of objective values will converge;

\item $\mathbf{P}_{t+1} - \mathbf{P}_{t} \rightarrow 0$, $\mathbf{B}_{t+1} - \mathbf{B}_{t} \rightarrow 0$,  $\mathbf{Z}_{t+1}^{(v)} - \mathbf{Z}_{t}^{(v)} \rightarrow 0 (1\le v \le V)$, and $\mathbf{F}_{t+1} - \mathbf{F}_{t} \rightarrow 0$;

\item The sequences $\{\mathbf{P}_t\}$, $\{\mathbf{B}_t\}$, $\{\mathbf{Z}_t^{(v)}\} (1\le v \le V)$ and $\{\mathbf{F}_t\}$ are bounded.
\end{enumerate}

\end{prop}

Assume the sequence $\{\{\mathbf{U}_{t}^{(v)}\}_{v=1}^V, \mathbf{W}_{t}\}$ generated by Algorithm \ref{alg:IML_BDR} is bounded, then we further have the following conclusion.

\begin{prop}\label{prop_solution}
Assume $\big\{\{\mathbf{U}_{t}^{(v)}\}_{v=1}^V, \mathbf{W}_{t}\big\}$ generated by Algorithm \ref{alg:IML_BDR} is bounded and $\mathbf{W}_{t+1} - \mathbf{W}_{t} \rightarrow 0$ and $\mathbf{U}_{t+1}^{(v)} - \mathbf{U}_{t}^{(v)} \rightarrow 0$ ($\forall v$), then there exists at least one subsequence that converges to a stationary point of Eq. (\ref{IML_BDR3}).
\end{prop}

The proofs of the above propositions are provided in the supplementary material.

\subsection{Computational Complexity}
Algorithm \ref{alg:IML_BDR} has six steps.
To solve $\{\mathbf{U}^{(v)}\}_{v=1}^V$, we need to calculate  $(\mathbf{W}\mathbf{W}^T)^{\dag}$ and multiply matrices, which cost $\mathcal{O}(r^3)$ and $\mathcal{O}(nrd^{(v)})$, respectively.
Then, the solution of $\mathbf{W}$ requires to solve a Sylvester equation.
The computational complexity of this step is $\mathcal{O}(r^3)$.
Solving $\mathbf{P}$ needs $\mathcal{O}(n^3)$, as the matrix inversion is involved.
To update $\mathbf{B}$, one needs to calculate $\mathbf{Q}$, which costs $\mathcal{O}(n^2k)$.
The optimal $\mathbf{F}$ is obtained by eigenvalue decomposition, spending $\mathcal{O}(n^3)$.
The updating of $\mathbf{Z}^{(v)}$ costs $\mathcal{O}(nrd^{(v)})$.
Hence, the overall time complexity of each iteration of Algorithm \ref{alg:IML_BDR} is $\mathcal{O}(2n^3+2r^3+n^2k+2ndr)$.
Recall that $r < \min(\{d^{(v)}\}, n))$ is the low-rank parameter, and is usually set as a small integer (such as $r=k$).
Thus, the dominate time complexity of each iteration is $\mathcal{O}(2n^3+n^2k+2ndr)$.

\begin{figure*}
  \centering
  \includegraphics[width=0.99\textwidth]{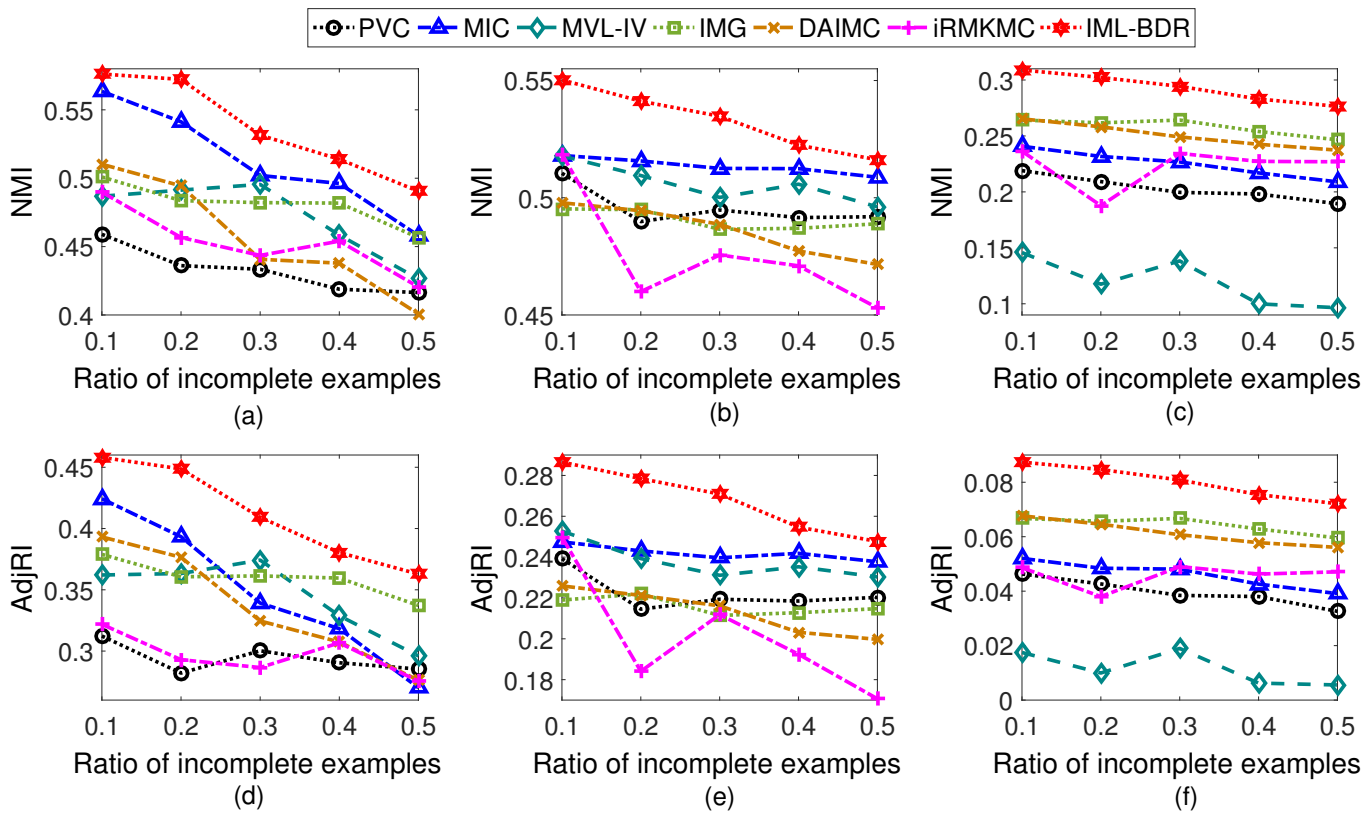}\\
  \caption{Comparisons on MSRC-v1, Yale and Corel5k (from left to right) w.r.t. NMI (1st row) and AdjRI (2rd row), respectively.}\label{fig_NMI_AdjRI}
\end{figure*}

\begin{table*}
  \centering
  \caption{NMI (mean(std.)) results of the compared methods as the incomplete ratio $m$ varies. The best results are highlighted in bold. }\label{Table_NMI}
 \begin{tabular}{|c|c|c|c|c|c|c| }
    \hline
Datasets	&$m$	&PVC  	& MVL-IV 	& IMG 	& DAIMC  	& IML-BDR   \\ \hline
\multirow{5}*{Caltech7 }&0.1		&0.403(0.058)	&0.602(0.027)	 &0.609(0.010)	&0.363(0.042)	&\textbf{0.684}(0.014)\\ \cline{2-7}
&0.2		&0.322(0.059)	&0.628(0.033)	&0.614(0.012)	 &0.281(0.034)	&\textbf{0.677}(0.020)\\ \cline{2-7}
&0.3		&0.235(0.029)	&0.637(0.023)	&0.556(0.066)	 &0.236(0.024)	&\textbf{0.666}(0.019)\\ \cline{2-7}
&0.4		&0.161(0.060)	&0.617(0.028)	&0.513(0.081)	 &0.199(0.018)	&\textbf{0.650}(0.018)\\ \cline{2-7}
&0.5		&0.127(0.049)	&0.610(0.028)	&0.420(0.024)	 &0.149(0.026)	&\textbf{0.636}(0.020)\\ \hline\hline
\multirow{5}*{Trecvid }&0.1		&0.143(0.051)	&0.236(0.036)	 &0.251(0.003)	&0.155(0.028)	&\textbf{0.281}(0.008)\\ \cline{2-7}
&0.2		&0.167(0.029)	&0.234(0.028)	&0.226(0.007)	 &0.142(0.009)	&\textbf{0.270}(0.010)\\ \cline{2-7}
&0.3		&0.113(0.043)	&0.213(0.044)	&0.210(0.010)	 &0.114(0.016)	&\textbf{0.261}(0.015)\\ \cline{2-7}
&0.4		&0.034(0.039)	&0.197(0.040)	&0.195(0.009)	 &0.090(0.021)	&\textbf{0.252}(0.012)\\ \cline{2-7}
&0.5		&0.009(0.002)	&0.168(0.042)	&0.175(0.005)	 &0.069(0.025)	&\textbf{0.235}(0.012)\\ \hline\hline
\multirow{5}*{PIE }&0.1		&0.141(0.005)	&0.140(0.015)	 &0.160(0.003)	&0.139(0.014)	&\textbf{0.172}(0.004)\\ \cline{2-7}
&0.2		&0.132(0.010)	&0.107(0.004)	&0.143(0.001)	 &0.119(0.001)	&\textbf{0.173}(0.008)\\ \cline{2-7}
&0.3		&0.115(0.005)	&0.110(0.007)	&0.132(0.003)	 &0.103(0.003)	&\textbf{0.175}(0.007)\\ \cline{2-7}
&0.4		&0.101(0.001)	&0.124(0.009)	&0.122(0.003)	 &0.090(0.000)	&\textbf{0.137}(0.012)\\ \cline{2-7}
&0.5		&0.096(0.002)	&0.092(0.005)	&0.106(0.005)	 &0.079(0.002)	&\textbf{0.114}(0.008)\\ \hline

  \end{tabular}
\end{table*}
\section{Experiment}\label{sec_exp}
In this section, we first verify the effectiveness of IML-BDR.
Then, we study how IML-BDR is affected by varying parameters. Finally, we show the advantages of the SOR-like algorithm in convergence.

\subsection{Data Preparation}
We perform experiments on six different real-world datasets.
They are Microsoft Research Cambridge Volume 1\footnote{https://www.microsoft.com/en-us/research/project/imageunderstanding/} (MSRC-v1),
Yale\footnote{http://vision.ucsd.edu/content/yale-face-database},
Corel5k\footnote{http://lear.inrialpes.fr/people/guillaumin/data.php},
Caltech101\footnote{http://www.vision.caltech.edu/Image\_Datasets/Caltech101/},
Trecvid2003\footnote{http://bigml.cs.tsinghua.edu.cn/$\thicksim$ningchen/data.htm} (Trecvid), and PIE\footnote{https://www.ri.cmu.edu/project/pie-database/}.
The detailed information of these datasets is described as follows.
We annotate the dimensionality of each view in the subsequent brackets.

\begin{itemize}
\item \textbf{MSRC-v1} has 240 images belonging to 8 classes. As the same in \cite{Cai2013MvKmeans}, we discard the background class, resulting in a dataset with 210 images in 7 classes. The SIFT (200) \cite{lowe2004SIFT} and LBP (256) \cite{ojala2002LBP} features are used.
 \item The \textbf{Yale} dataset is a face image database. There are 165 grayscale images of 15 individuals, and each subject has 11 images. SIFT (50), GIST (512) \cite{oliva2001GIST}, and LBP (256) features are extracted for experiments.

 \item The \textbf{Corel5k} contains 4,999 images from 50 categories. We use 3 kinds of pre-extracted features by M. Guillaumin et al. \cite{Guillaumin2009TagProp} for experiments. The features are GIST (512), DenseSIFT (1000), and DenseHue (100).

 \item Caltech101 is a collection of images for object recognition. It consists of 101 kinds of objects.
      Following \cite{Cai2013MvKmeans}, we use a subset that contains 441 images of 7 classes for experiments.
      The subset is referred to as \textbf{Caltech7}. SIFT (200), SURF (200) \cite{bay2006SURF}, and LBP (256) features are extracted.

  \item The \textbf{Trecvid} dataset consists of 1078 video shots belonging to 5 categories. Each shot has two kinds of feature representations, i.e., the text feature (1894) and the HSV color histogram (165) extracted from the associated keyframe.

  \item The \textbf{PIE} dataset is a subset of the CMU PIE face database. This subset is composed of the images of five near frontal poses (C05, C07, C09, C27, C29) and all the images under different illuminations and expressions. The resultant dataset has 11,554 samples belonging to 68 categories. SIFT (50) and LBP (256) features are extracted.

\end{itemize}

We prepared datasets for the missing-view setting and the incomplete-view setting, respectively.
In the missing-view setting, the datasets used are MSRC-v1, Yale and Corel5k.
Since the datasets are originally complete, we construct the missing-view setting as follows.
We randomly select $m$ percent (10\% to 50\%) examples and randomly discard one view from each example.
In the incomplete-view setting, Caltech7, Trecvid and PIE are used.
In this setting, datasets have both the missing views and missing variables, which are prepared as follows.
The first step is the same as the missing-view setting, i.e., $m$ percent (10\% to 50\%) examples are randomly selected with one random view being removed for each example.
Then, on each view, $m$ percent (10\% to 50\%) entries are randomly removed from the matrix formed by the rest examples.
For both settings, to avoid the inaccuracy brought by randomness, the construction process is repeated 5 times for Corel5k and PIE and 10 times for the rest datasets.

\subsection{Baselines and Evaluation Metrics}
Firstly, we evaluate the performance of different IML methods by conducting clustering on the learned embedding matrix, in the missing-view setting and the incomplete-view setting respectively.
In the missing view setting, IML-BDR is compared with PVC \cite{Li2014PVC}, MIC \cite{shao2015MIC}, MVL-IV \cite{xu2015MVL}, IMG \cite{Zhao2016IMG}, DAIMC \cite{Hu2018DAIMC}, and iRMKMC which is adapted from RMKMC \cite{Cai2013MvKmeans} in Sec. \ref{subsec_pRMKMC}.
The iMCL adapted from MCL \cite{Guan2015MCL} in Sec. \ref{subsec_pRMKMC} is not included into the comparison, since it is a semi-supervised method while all the above mentioned algorithms are unsupervised.
In the incomplete-view setting, since MIC and iRMKMC are not applicable, IML-BDR is compared with MVL-IV and the extended PVC (Eq. (\ref{eq_multiPVC2})), IMG (Eq. (\ref{eq_multiIMG2})) and DAIMC (Eq. (\ref{eq_DAIMC3})).

Then, we compare IML-BDR with two completion methods in terms of reconstruction ability in the incomplete-view setting.
The two completion methods are Robust Principle Component Analysis (RPCA) \cite{Lin2009ALM} and Robust Rank-$k$ Matrix Completion (RRMC) \cite{Huang2013rank_MC}.
After completion, we run the complete Robust Auto-Weighted Multi-View Clustering (RAMC) \cite{ren2018RAMC} method on the recovered matrices, and compare the clustering performance with that of IML-BDR.

As the previous works \cite{Li2014PVC,Zhao2016IMG} do, the dimensionality of the embedding in all compared IML methods is set to be the number of clusters, i.e., $r = k$.
The regularization parameter $\alpha$ in PVC is tuned from $\{10^{-4}, 10^{-3}, 10^{-2}, 10^{-1}\}$.
For MIC, The co-regularization parameters $\{\alpha_i\}$ of MIC are set to 0.01, and the robust parameters $\{\beta_i\}$ are all tuned from $\{10^{-3}, 10^{-2}, 10^{-1}\}$ according to the parameter study in the original paper.
As suggested by the authors \cite{Zhao2016IMG}, $\gamma$ in IMG are fixed as 100, and $\alpha$ and $\beta$ of IMG are selected from $\{0.001, 0.01, 0.1\}$ and $\{0.1, 1, 10\}$, respectively.
For DAIMC, both $\alpha$ and $\beta$ are tuned from $\{0.1, 1, 10\}$.
The $\gamma$ ($\ge 1$) in iRMKMC is chosen from $\{2, 3, 5\}$.
There is no parameter need to be tuned in RPCA and RAMC.
For RRMC, the rank of the recovered matrix $r$ is tuned from $\{k, 50, 100\}$, where $k$ is the number of clusters.
For IML-BDR, $\beta$ is set as $10^{4}$, $\alpha$ and $\gamma$ are tuned from $\{10^{-2}, 1, 10^{2}\}$ and $\{1, 10, 10^{2}\}$, respectively.

For the fairness of comparisons, except for iRMKMC, all the other compared IML methods conduct $k$-means clustering on the learned embedding matrix to obtain the final partitions.
The normalized mutual information (NMI) and the adjusted rand index (AdjRI) are utilized for clustering performance evaluation.
The higher the two metrics' scores are, the better the clustering performance is.
The $k$-means clustering is repeated 20 times, and the mean value of NMI (or AdjRI) is used as the result for each independently constructed missing-view or incomplete-view repetition.
Finally, the average performance over all the repetitions is presented.

As for the reconstruction error, we use the widely used root mean square error (RMSE) for comparison. Denote the recovered matrix and the ground truth matrix as $\mathbf{M}$ and $\mathbf{M}^* \in \mathbb{R}^{m\times n}$ respectively, then RMSE is defined as $\frac{1}{\sqrt{mn}}\|\mathbf{M}^* - \mathbf{M}\|$.
The smaller RMSE is, the better the recovered matrix is.

\begin{table*}
  \centering
  \caption{AdjRI (mean(std.)) results of the compared methods as the incomplete ratio $m$ varies. The best results are highlighted in bold.  }\label{Table_AdjRI}
  \begin{tabular}{|c|c|c|c|c|c|c|}
    \hline
Datasets	&$m$	&PVC  	& MVL-IV 	& IMG 	& DAIMC  	& IML-BDR   \\ \hline
\multirow{5}*{Caltech7 }&0.1		&0.317(0.068)	&0.507(0.043)	 &0.523(0.017)	&0.298(0.047)	&\textbf{0.602}(0.034)\\ \cline{2-7}
&0.2		&0.240(0.049)	&0.530(0.043)	&0.529(0.022)	 &0.222(0.035)	&\textbf{0.590}(0.048)\\ \cline{2-7}
&0.3		&0.179(0.032)	&0.555(0.037)	&0.470(0.067)	 &0.184(0.022)	&\textbf{0.581}(0.037)\\ \cline{2-7}
&0.4		&0.110(0.053)	&0.524(0.041)	&0.437(0.076)	 &0.137(0.019)	&\textbf{0.563}(0.027)\\ \cline{2-7}
&0.5		&0.072(0.035)	&0.521(0.032)	&0.340(0.027)	 &0.081(0.020)	&\textbf{0.560}(0.022)\\ \hline\hline
\multirow{5}*{Trecvid }&0.1		&0.109(0.047)	&0.184(0.044)	 &0.209(0.006)	&0.100(0.045)	&\textbf{0.237}(0.006)\\ \cline{2-7}
&0.2		&0.147(0.050)	&0.193(0.035)	&0.182(0.009)	 &0.129(0.019)	&\textbf{0.227}(0.009)\\ \cline{2-7}
&0.3		&0.091(0.041)	&0.172(0.045)	&0.160(0.014)	 &0.094(0.028)	&\textbf{0.220}(0.011)\\ \cline{2-7}
&0.4		&0.025(0.032)	&0.142(0.046)	&0.138(0.015)	 &0.064(0.025)	&\textbf{0.210}(0.009)\\ \cline{2-7}
&0.5		&0.003(0.005)	&0.129(0.040)	&0.107(0.005)	 &0.056(0.018)	&\textbf{0.191}(0.010)\\ \hline\hline
\multirow{5}*{PIE }&0.1		&0.013(0.001)	&0.021(0.003)	 &0.016(0.000)	&0.015(0.004)	&\textbf{0.022}(0.001)\\ \cline{2-7}
&0.2		&0.012(0.002)	&0.013(0.001)	&0.013(0.000)	 &0.011(0.001)	&\textbf{0.023(}0.003)\\ \cline{2-7}
&0.3		&0.009(0.001)	&0.015(0.001)	&0.010(0.001)	 &0.008(0.000)	&\textbf{0.023}(0.001)\\ \cline{2-7}
&0.4		&0.007(0.000)	&\textbf{0.019}(0.002)	&0.008(0.000)	 &0.006(0.000)	&0.016(0.004)\\ \cline{2-7}
&0.5		&0.004(0.001)	&\textbf{0.011}(0.001)	&0.006(0.000)	 &0.005(0.000)	&0.010(0.001)\\ \hline
  \end{tabular}
\end{table*}

\begin{table*}
  \centering
  \caption{Reconstruction errors compared with completion methods RPCA and RRMC and clustering results compared with running RAMC on the recovered data matrices by RPCA and RRMC, respectively. $m$ denotes the incomplete ratio. The best results are highlighted in bold.  }\label{Table_MC}
  \begin{tabular}{|c|c|c|c|c|c|c|c|c|c|c|}
    \hline
\multirow{2}*{Datasets} 	& \multirow{2}*{$m$} 	& \multicolumn{3}{c|}{RPCA(+RAMC)} 	& \multicolumn{3}{c|}{RRMC(+RAMC)} 	 & \multicolumn{3}{c|}{IML-BDR}  \\ \cline{3-11}
 	& 	& RMSE $\downarrow$ 	& NMI $\uparrow$ 	& AdjRI $\uparrow$ 	 & RMSE $\downarrow$	& NMI $\uparrow$  & AdjRI  $\uparrow$  	& RMSE $\downarrow$	& NMI  $\uparrow$ 	& AdjRI  $\uparrow$ \\ \hline
\multirow{6}*{Caltech7 }& 0 & - & .592(.049) & .334(.084) & -  & .592(.049) & .334(.084) & - & \textbf{.703}(.044) & \textbf{.644}(.078) \\ \cline{2-11}
&0.1		&.0477(.0001)	&.103(.009)	&.060(.021)	&.0955(.0013)	 &.397(.060)	&.143(.085)	&\textbf{.0331}(.0001)	&\textbf{.684}(.014)	 &\textbf{.602}(.034)\\ \cline{2-11}
&0.2		&.0478(.0001)	&.086(.009)	&.043(.014)	&.1258(.0021)	 &.372(.074)	&.108(.070)	&\textbf{.0333}(.0001)	&\textbf{.677}(.020)	 &\textbf{.590}(.048)\\ \cline{2-11}
&0.3		&.0478(.0002)	&.083(.009)	&.041(.019)	&.1526(.0030)	 &.325(.081)	&.088(.043)	&\textbf{.0335}(.0001)	&\textbf{.666}(.019)	 &\textbf{.581}(.037)\\ \cline{2-11}
&0.4		&.0480(.0001)	&.074(.012)	&.039(.015)	&.1783(.0048)	 &.328(.060)	&.089(.049)	&\textbf{.0337}(.0001)	&\textbf{.650}(.018)	 &\textbf{.563}(.027)\\ \cline{2-11}
&0.5		&.0481(.0002)	&.061(.009)	&.034(.014)	&.1962(.0033)	 &.302(.082)	&.082(.054)	&\textbf{.0342}(.0001)	&\textbf{.636}(.020)	 &\textbf{.560}(.022)\\ \hline\hline
\multirow{6}*{Trecvid }& 0 & - & .223(.007) & .062(.004) & - &  .223(.007) & .062(.004) & - & \textbf{.274}(.002) &\textbf{ .232}(.008) \\ \cline{2-11}
&0.1		&.1730(.0007)	&.086(.018)	&.055(.013)	&.1141(.0034)	 &.191(.034)	&.045(.014)	&\textbf{.0252}(.0000)	&\textbf{.281}(.008)	 &\textbf{.237}(.006)\\ \cline{2-11}
&0.2		&.1739(.0010)	&.080(.009)	&.055(.012)	&.1589(.0061)	 &.165(.032)	&.049(.008)	&\textbf{.0253}(.0000)	&\textbf{.270}(.010)	 &.\textbf{227}(.009)\\ \cline{2-11}
&0.3		&.1758(.0011)	&.084(.010)	&.058(.010)	&.1936(.0040)	 &.171(.042)	&.047(.007)	&\textbf{.0254}(.0000)	&\textbf{.261}(.015)	 &\textbf{.220}(.011)\\ \cline{2-11}
&0.4		&.1761(.0012)	&.076(.014)	&.047(.009)	&.2212(.0047)	 &.118(.021)	&.045(.014)	&\textbf{.0257}(.0001)	&\textbf{.252}(.012)	 &\textbf{.210}(.009)\\ \cline{2-11}
&0.5		&.1777(.0015)	&.069(.015)	&.050(.015)	&.2467(.0031)	 &.114(.031)	&.037(.011)	&\textbf{.0258}(.0001)	&\textbf{.235}(.012)	 &\textbf{.191}(.010)\\ \hline\hline
\multirow{5}*{PIE} & 0 & - & .186(.002) & .0004(.0000) & - & .186(.002) & .0004(.0000) & -& \textbf{.208}(.003) & \textbf{.029}(.001) \\ \cline{2-11}
&0.1		&.0456(.0000)	&.065(.001)	&.002(.000)	&.1083(.0011)	 &\textbf{.178}(.003)	&.0004(.0001)	&\textbf{.0260}(.0008)	 &.172(.004)	 &\textbf{.022}(.001)\\ \cline{2-11}
&0.2		&.0458(.0000)	&.065(.002)	&.002(.000)	&.1526(.0012)	 &.132(.025)	&.0003(.0000)	&\textbf{.0315}(.0037)	&\textbf{.173}(.008)	 &\textbf{.023}(.003)\\ \cline{2-11}
&0.3		&.0460(.0000)	&.059(.000)	&.002(.000)	&.1869(.0024)	 &.136(.034)	&.0002(.0000)	&\textbf{.0306}(.0010)	&\textbf{.175}(.007)	 &\textbf{.023}(.001)\\ \cline{2-11}
&0.4		&.0463(.0000)	&.060(.002)	&.001(.000)	&.2162(.0015)	 &.124(.035)	&.0002(.0000)	&.\textbf{0359}(.0037)	&\textbf{.137}(.012)	 &\textbf{.016}(.004)\\ \cline{2-11}
&0.5		&.0468(.0000)	&.057(.002)	&.001(.000)	&.2432(.0011)	 &\textbf{.114}(.030)	&.0001(.0000)	&\textbf{.0449}(.0030)	 &\textbf{.114}(.008)	 &\textbf{.010}(.001)\\ \hline
  \end{tabular}
\end{table*}

\subsection{Comparison Results}
Fig. \ref{fig_NMI_AdjRI} and Tables \ref{Table_NMI} - \ref{Table_AdjRI} show the clustering results in the missing-view setting and incomplete-view setting, respectively.
According to these results, we have the following observations.

 As the ratio of incomplete examples increases, it can be seen that the performance of all the compared methods is degenerated in most cases.
This is consistent with the intuition.
Some exceptions exist may be because of the random construction of incomplete datasets.

MIC consistently outperforms PVC on the three datasets in the missing-view setting. Both of them are based on NMF. What is different is that, PVC directly learns a common embedding matrix and imposes sparse $\ell_{1,1}$ regularization onto it, while MIC learns the common embedding by pushing each view's $\ell_{2,1}$-norm regularized embedding matrix towards a common consensus with a weighting scheme. It is probably that the weighting scheme and $\ell_{2,1}$ regularization make MIC more robust to missing views than PVC.

Based on semi-NMF, DAIMC has more fluctuant performance than PVC and MIC over different datasets. This might be because that simultaneously learning a common latent embedding matrix and establishing a consensus basis matrix limit DAIMC's representation ability for various datasets.

Although MVL-IV does not employ any regularization on the common embedding or the basis matrices, sometimes its performance is not bad due to its flexibility.

Compared with its performance in the missing-view setting, IMG suffers from more performance degeneration as the incomplete ratio increases in the incomplete-view setting.
The possible reason is that the Laplacian graph regularization is not robust to random missing variables.

Despite the performance of iRMKMC is not very good, it can provide a means to process multi-view data with missing views, which is an alternative when there is no time to design new IML algorithms.
This verifies that the proper use of the proposed JELLA framework can improve the efficiency of processing incomplete multi-view data.

In the missing-view setting, the proposed IML-BDR consistently outperforms the existing methods in terms of both metrics.
In the incomplete-view setting, it also achieves the highest NMI or AdjRI score in most cases.
These results verify the effectiveness of IML-BDR.
Compared with the baselines, the utilization of the BDR term enables IML-BDR to focus more on exploring the underlying clustering structures and preserve the representation capability in case of missing views and missing variables.
As a result, the embedding matrix learned by IML-BDR is more discriminative, which further leads to the clustering performance improvements of IML-BDR.

Table \ref{Table_MC} displays the comparison results of the reconstruction error between IML-BDR and the two completion methods: RPCA and RRMC, in the incomplete-view setting.
Since RPCA and RRMC are single-view methods, for each datasets, we first concatenate the incomplete data matrices from different views into a big matrix and then apply the completion methods.
As shown in Table \ref{Table_MC}, as the incomplete ratio increases, the reconstruction error becomes larger in most cases.
RPCA gets considerable results on Caltech7 and PIE.
RRMC is founded to be not good at restoring the missing views and missing variables on all datasets.
IML-BDR achieves the smallest RMSE values for different datasets for all incomplete ratios.
Compared with RPCA and RRMC, the advantage of IML-BDR with respect to recovery ability mainly results from the following factor.
That is, IML-BDR employs the BDR term to exploit the useful information from different views in a more elegant way rather than using the simple concatenation.

Table \ref{Table_MC} also presents the clustering results of running the RAMC \cite{ren2018RAMC}, which is a complete multi-view clustering method, on the recovered data matrices by RPCA and RRMC.
The corresponding methods are denoted as ``RPCA + RAMC'' and ``RRMC + RAMC'', respectively.
When the incomplete ratio is 0, i.e., when the input data matrices are complete, RAMC can be directly applied.
As either RPCA or RRMC is not implemented, the clustering results for ``RPCA + RAMC'' and ``RRMC + RAMC'' are the same.
The results show that IML-BDR outperforms the state-of-the-art RAMC on all three datasets.
When the data is incomplete, first filling the missing values by RPCA or RRMC and then applying RRMC degenerates the clustering performance in comparison with the complete case.
In different missing ratios, the clustering results of ``RPCA + RAMC'' and ``RRMC + RAMC'' are worse than those of IML-BDR in most cases.
These results show that IML-BDR is effective bor both complete and incomplete multi-view data.

\begin{figure}
\centering
\subfigure[]{
\includegraphics[width=0.22\textwidth]{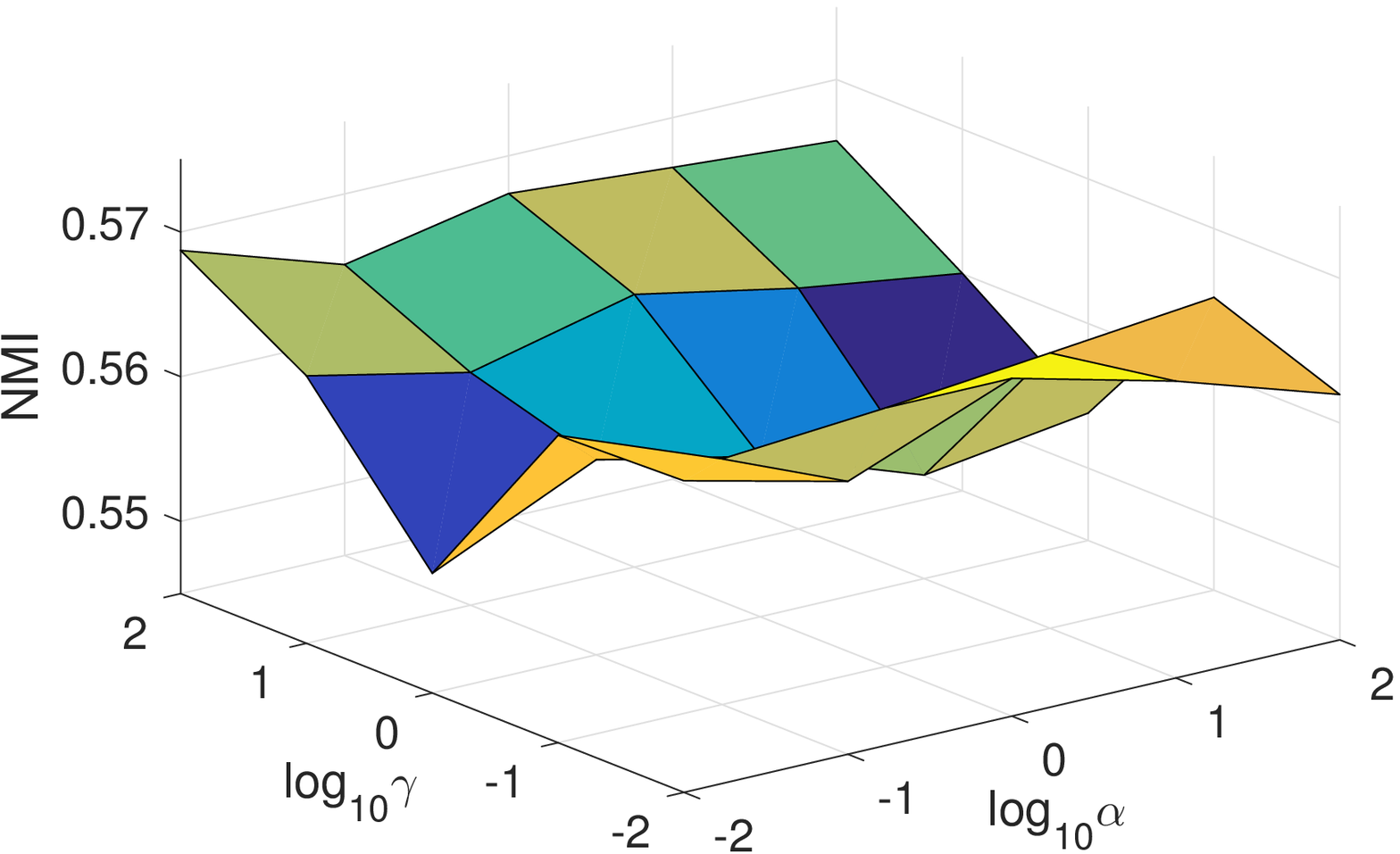}}
\subfigure[]{
\includegraphics[width=0.22\textwidth]{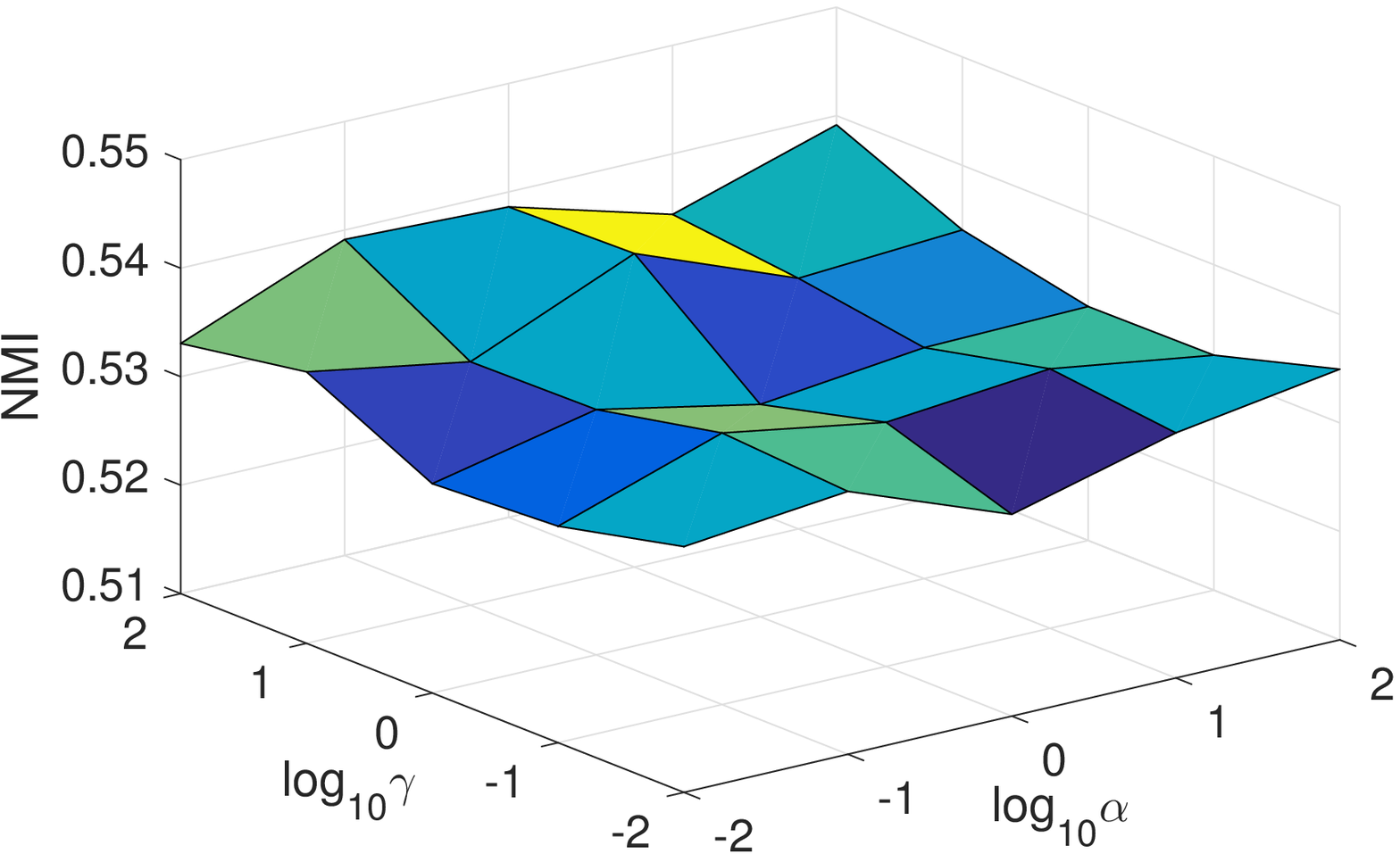}}
\centering
\caption{Effect of parameters ($\alpha$ and $\gamma$) evaluated by NMI on (a) MSRC-v1 and (b) Yale.}
\label{Fig.ParaNMI}
\end{figure}

\begin{figure}
\centering
\subfigure[]{
\includegraphics[width=0.22\textwidth]{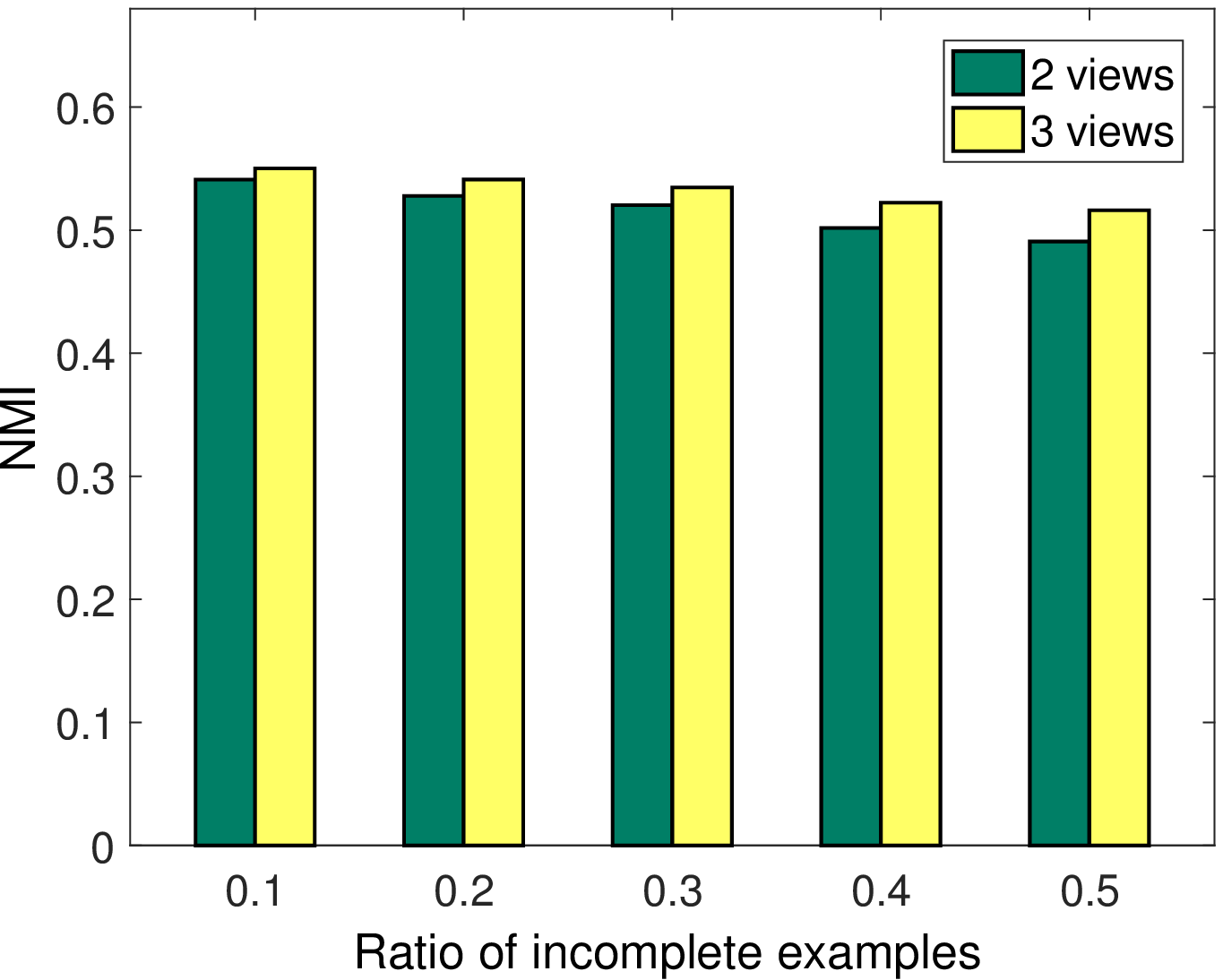}}
\subfigure[]{
\includegraphics[width=0.22\textwidth]{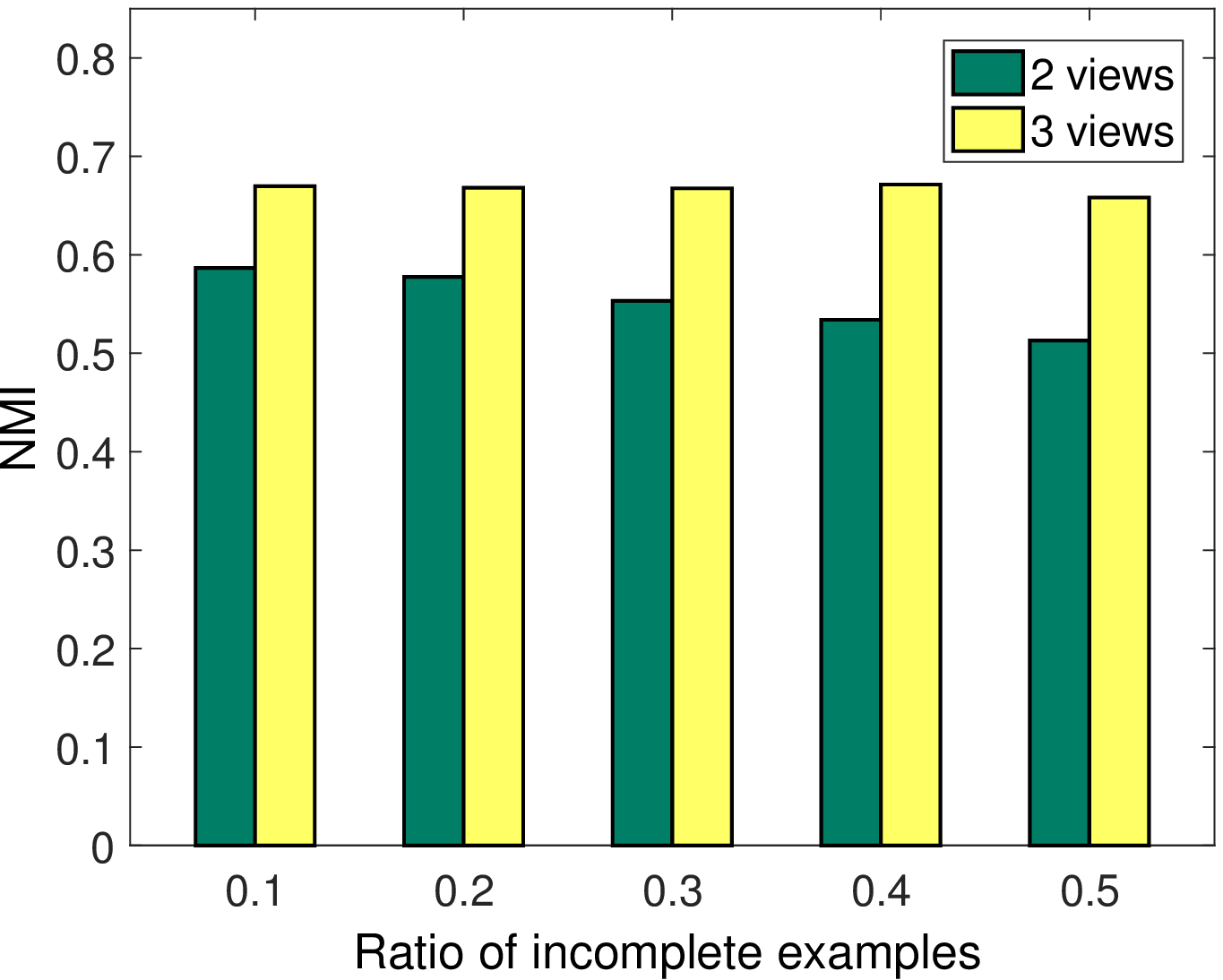}}
\centering
\caption{Effect of view numbers. (a) Yale. (b) Caltech7.}
\label{Fig.viewComp}
\end{figure}

\begin{figure}
\centering
\subfigure[]{
\includegraphics[width=0.22\textwidth]{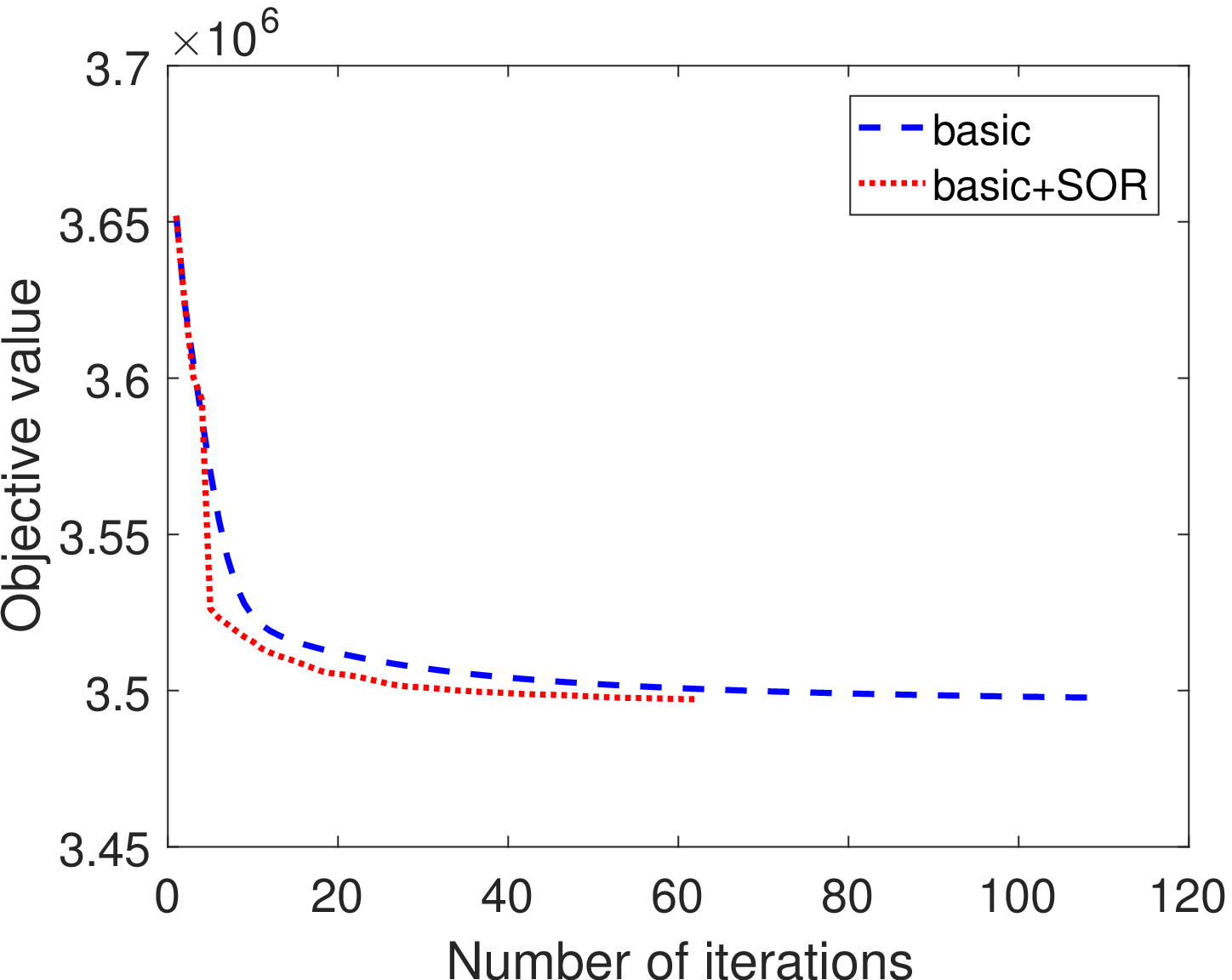}}
\subfigure[]{
\includegraphics[width=0.22\textwidth]{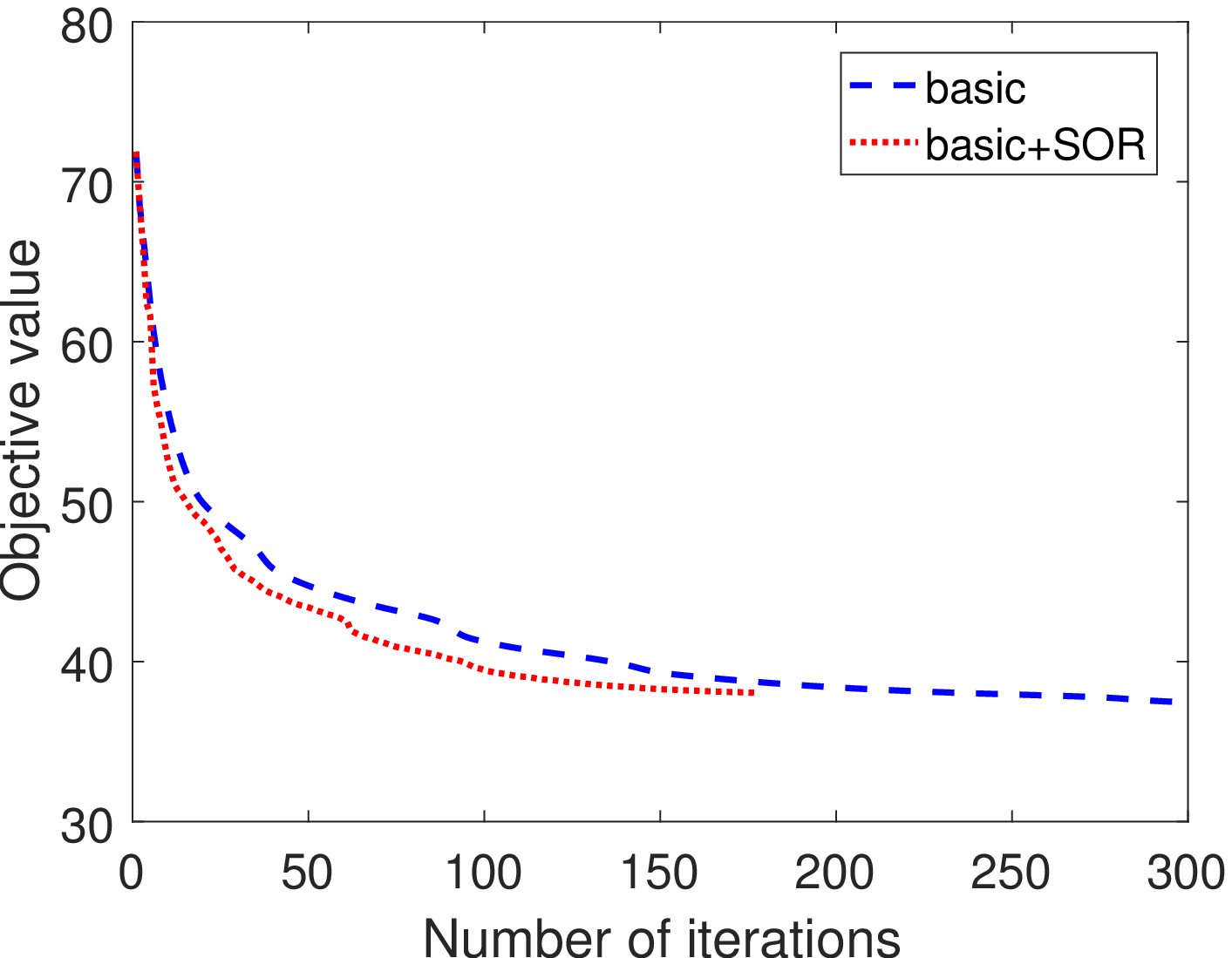}}
\centering
\caption{Objective curves of IML-BDR on (a) Caltech7 and (b) Yale.}
\label{Fig.objCurve}
\end{figure}

\subsection{Parameter Study}
In this subsection, how IML-BDR is affected by varying parameters ($\alpha$ and $\gamma$) is studied.
We vary the value of both parameters within $\{ 10^{-2}, 10^{-1}, 1, 10^{1}, 10^{2}\}$.
Without loss of generality, the experiments are performed when the incomplete ratio is 0.2 in the missing-view setting.
The NMI results on MSRC-v1 and Yale are shown in Fig. \ref{Fig.ParaNMI}.
It can be seen that the performance of IML-BDR is not much changed as the parameter varies.

Besides, we also test how IML-BDR's performance is affected by the varying number of views.
IML-BDR is tested on Yale and Caltech7 in the  missing-view setting and incomplete-view setting, respectively.
The results are shown in Fig. \ref{Fig.viewComp}. Only one of the three combinations in the 2-view case is shown, since they have similar results.
It can be seen that IML-BDR performs better with more views.
The results are as expected: with more views, more information can be provided from the other views for the missing values, leading to more accurate learning.

\subsection{Convergence}
In this subsection, the convergence behavior of Algorithm \ref{alg:IML_BDR} is tested.
We use ``basic'' to denote the basic steps to solve IML-BDR in Eqs. (\ref{eq_sub1}) - (\ref{eq_sub6}), and ``basic + SOR'' means the SOR-like optimization procedure in Algorithm \ref{alg:IML_BDR}.
The algorithm is deemed to be converged, if the relative variation in objective values between two consecutive iterations is less than $10^{-4}$.
We show the objective curves of IML-BDR on Yale and  Catech7 in Fig. \ref{Fig.objCurve}.
As we can see, both strategies converge after a number of iterations, and the ``basic + SOR'' strategy converges faster than the ``basic'' one.

In addition, we report the training time measured in seconds and the corresponding number of iterations in Table \ref{tab_SOR} for the two optimization strategies.
It can be seen that the utilization of the SOR technique helps to shorten the training time and reduce the number of iterations for convergence.

\begin{table}
\centering
\caption{Training time measured in seconds (number of iterations) comparison in optimizing IML-BDR with different strategies on two datasets.}\label{tab_SOR}
\scriptsize{\begin{tabular}{|c|c|c|}
  \hline
  \multirow{2}*{Strategy} & \multicolumn{2}{c|}{Datasets} \\ \cline{2-3}
  & Caltech7 & Trecvid   \\ \hline
  basic & 7.877(109) & 85.931(173)   \\ \hline
  basic+SOR & 5.183(62) & 47.605(84)   \\
  \hline
\end{tabular}}
\end{table}

\section{Conclusion}\label{sec_summary}
In this paper, we propose the JELLA framework to provide a unified perspective for understanding several existing IML methods.
With the guidance of this framework, some linear transformation based complete multi-view methods can be adapted to IML directly.
This bridges the gap between complete multi-view learning and IML, and is of practical significance in improving the efficiency of dealing with incomplete multi-view data.
Moreover, this framework can also provide guidance for developing new algorithms.
As shown in this paper, within the framework, we propose the IML-BDR algorithm from the perspective of subspace clustering.
IML-BDR pursues the block diagonal property to obtain better subspace clustering. An SOR-like optimization algorithm with guaranteed convergence is developed to solve IML-BDR.
Experimental results on various datasets validate the effectiveness of IML-BDR.

\bibliographystyle{IEEEtran}

\ifCLASSOPTIONcaptionsoff
  \newpage
\fi

\end{document}